%% file: main.tex
\documentclass[journal]{IEEEtran}
\input{ieee_preamble.tex}
\usepackage[hidelinks,hypertexnames=false]{hyperref}
\input{posting_notice.tex}
\let\citep\cite
\let\citet\cite
\let\citealp\cite
\newif\ifshamisacombinedbuild
\shamisacombinedbuildtrue
\hypersetup{pdftitle={\shamisatitle},pdfauthor={Mahdi Naseri and Zhou Wang},pdfsubject={Author-accepted manuscript; DOI: 10.1109/TIP.2026.3744700}}
\title{\shamisatitle}
\author{\shamisaauthors\thanks{\shamisapostingnotice}}
\begin{document}
\maketitle

\input{1_main}
\section*{Supplementary Material}
The appendices referenced throughout this manuscript are provided in the accompanying supplementary material.
\input{main.bbl}

\clearpage
\setcounter{page}{1}
\setcounter{figure}{6}
\setcounter{table}{3}
\setcounter{equation}{19}
\markboth{Supplementary Material for \shamisashorttitle}{Naseri and Wang: \shamisashorttitle}
\begin{center}
{\normalsize\bfseries Supplementary Material for ``\shamisatitle''\par}
\vspace{0.25em}
{\small Mahdi Naseri and Zhou Wang\par}
\end{center}
\vspace{0.5em}
\makeatletter
\newcommand{\shamisasuppcite}[1]{\begingroup
  \def\shamisacitelist{}%
  \@for\shamisakey:=#1\do{%
    \ifx\shamisacitelist\@empty
      \edef\shamisacitelist{supp:\shamisakey}%
    \else
      \edef\shamisacitelist{\shamisacitelist,supp:\shamisakey}%
    \fi}%
  \expandafter\cite\expandafter{\shamisacitelist}\endgroup}
\makeatother
\let\citep\shamisasuppcite
\let\citet\shamisasuppcite
\let\citealp\shamisasuppcite
\input{1_supplementary}
\input{supplement_references.bbl}

\end{document}

%% file: ieee_preamble.tex
\usepackage[T1]{fontenc}
\usepackage{graphicx}
\usepackage{booktabs}
\usepackage{multirow}
\usepackage{mathtools}
\usepackage{mathrsfs}
\usepackage{rotating}
\usepackage[caption=false,font=footnotesize]{subfig}
\usepackage{tikz}
\usepackage{xcolor}
\usepackage{url}
\usepackage{cite}
\usepackage{algorithm}
\usepackage{algorithmic}

\input{math_commands.tex}

\newcommand{\srcc}{SRCC}
\newcommand{\plcc}{PLCC}

\newcommand{\live}{LIVE}
\newcommand{\csiq}{CSIQ}
\newcommand{\tid}{TID2013}
\newcommand{\kadid}{KADID-10K}

\newlength{\figup}
\newlength{\figdown}
\newlength{\tabdown}
\providecommand{\citep}[1]{\cite{#1}}
\providecommand{\citet}[1]{\cite{#1}}
\providecommand{\citealp}[1]{\cite{#1}}

\newcommand{\shamisaauthors}{Mahdi Naseri and Zhou Wang%
\thanks{Mahdi Naseri and Zhou Wang are with the Department of Electrical and Computer Engineering, University of Waterloo, Waterloo, ON N2L 3G1, Canada (e-mail: mahdi.naseri@uwaterloo.ca; zhou.wang@uwaterloo.ca). Corresponding author: Mahdi Naseri.}}

\newcommand{\shamisatitle}{SHAMISA: SHAped Modeling of Implicit Structural Associations for Self-supervised No-Reference Image Quality Assessment}
\newcommand{\shamisashorttitle}{SHAMISA}
\newcommand{\shamisakeywords}{blind image quality assessment, self-supervised learning, graph-based representation learning, distortion modeling, perceptual quality prediction}

%% file: math_commands.tex
\usepackage{amsmath,amsfonts,bm}

\def\eqref#1{equation~\ref{#1}}

\def\1{\bm{1}}

\def\mA{{\bm{A}}}

\def\mC{{\bm{C}}}

\def\mG{{\bm{G}}}
\def\mH{{\bm{H}}}

\def\mR{{\bm{R}}}
\def\mS{{\bm{S}}}
\def\mT{{\bm{T}}}

\def\mX{{\bm{X}}}

\def\mZ{{\bm{Z}}}

\DeclareMathAlphabet{\mathsfit}{\encodingdefault}{\sfdefault}{m}{sl}
\SetMathAlphabet{\mathsfit}{bold}{\encodingdefault}{\sfdefault}{bx}{n}

\newcommand{\R}{\mathbb{R}}

\newcommand{\Cov}{\mathrm{Cov}}


%% file: posting_notice.tex
\newcommand{\shamisapostingnotice}{Author-accepted manuscript of the article published in
\emph{IEEE Transactions on Image Processing}, vol.~35, pp.~11216--11227, 2026.
The final published version is available at
\href{https://doi.org/10.1109/TIP.2026.3744700}{10.1109/TIP.2026.3744700}.
\copyright\ 2026 IEEE. Personal use of this material is permitted.
Permission from IEEE must be obtained for all other uses, in any current or
future media, including reprinting/republishing this material for advertising
or promotional purposes, creating new collective works, for resale or
redistribution to servers or lists, or reuse of any copyrighted component
of this work in other works.}

%% file: 1_main.tex
\begin{abstract}
    No-Reference Image Quality Assessment (NR-IQA) aims to estimate perceptual quality without access to a pristine reference. Learning an NR-IQA model faces a fundamental bottleneck because it requires many costly human perceptual labels. We propose SHAMISA, a non-contrastive self-supervised framework that learns from unlabeled distorted images through explicitly structured relational supervision. Unlike prior methods that impose rigid binary similarity constraints, SHAMISA shapes soft, controllable implicit structural associations that are both distortion-aware and content-sensitive. A key innovation is its compositional distortion engine, which generates an uncountable family of degradations from continuous parameter spaces and groups the resulting images so that exactly one distortion factor varies at a time. This construction enables fine-grained control over representational similarity during training. Images with shared distortion patterns are pulled together in the embedding space, while severity variations produce structured shifts. Dual-source relation graphs integrate known degradation profiles with emergent structural affinities throughout pre-training. The learned encoder is then frozen and paired with a linear regressor. Extensive experiments on synthetic, authentic, and cross-dataset NR-IQA benchmarks demonstrate strong overall performance, improved transfer in the reported cross-dataset comparisons, and favorable qualitative robustness diagnostics. Self-supervised pre-training requires no IQA opinion-score annotations or contrastive loss.
\end{abstract}

\newcommand{\suppAppArchitectures}{Supplementary Appendix A}
\newcommand{\suppAppDatasets}{Supplementary Appendix D}
\newcommand{\suppAppFR}{Supplementary Appendix E}
\newcommand{\suppAppCross}{Supplementary Appendix F}
\newcommand{\suppAppTransfer}{Supplementary Appendix G}
\newcommand{\suppApptsne}{Supplementary Appendix H}
\newcommand{\suppAppumap}{Supplementary Appendix I}
\newcommand{\suppAppgmad}{Supplementary Appendix J}
\newcommand{\suppAppDynamics}{Supplementary Appendix L}
\newcommand{\suppAppHPSens}{Supplementary Appendix M}
\newcommand{\suppTabFR}{Supplementary Table VI}
\newcommand{\suppTabTransfer}{Supplementary Table VII}
\newcommand{\suppTabGMAD}{Supplementary Table VIII}
\newcommand{\suppTabAbl}{Supplementary Table IX}
\newcommand{\suppTabUncertainty}{Supplementary Table X}

\begin{IEEEkeywords}
    \shamisakeywords
\end{IEEEkeywords}

\section{Introduction}
\IEEEPARstart{I}{mage} Quality Assessment (IQA) aims to estimate perceptual image quality in line with human opinion. The No-Reference setting (NR-IQA) is especially challenging, as it must operate without reference images or distortion labels. NR-IQA is critical for real-world tasks such as perceptual enhancement \citep{wang2021real}, image captioning \citep{chiu2020assessing}, and streaming optimization \citep{agnolucci2023perceptual}. However, modeling perceptual quality is difficult due to the complex interplay between distortion types and content \citep{Bovik2013AutomaticPO, 4775883}. Supervised approaches \citep{zhang2018blind, su2020blindly, golestaneh2022no} rely on extensive human annotations, with KADID-10K \citep{kadid10k} alone requiring over 300,000 subjective ratings, making them expensive and hard to scale. To address this, recent efforts turn to self-supervised learning (SSL), using unlabeled distorted images to learn quality-aware representations \citep{zhao2023quality, reiqa}, typically via contrastive objectives tailored to either content or degradation similarity.

Classical SSL methods like SimCLR \citep{chen2020simple} and MoCo \citep{he2020momentum}, developed for classification, learn content-centric and distortion-invariant features. Such representations misalign with NR-IQA, where both content and degradation must be modeled. To bridge this gap, contrastive SSL-IQA methods introduce domain-specific objectives but can suffer from sampling bias because sampled negatives may include semantically related false negatives \citep{bielak2021graph,chuang2020debiased,chen2022incremental}.
CONTRIQUE \citep{madhusudana2022image} groups images by distortion type and severity, ignoring content. Re-IQA \citep{reiqa}, by contrast, uses overlapping crops of the same image to promote content-aware alignment. While each mode captures useful relations, similarity is enforced either across content or across distortions, but not both. This leads to scattered embeddings for similarly degraded images with different content \citep{arniqa}. ARNIQA \citep{arniqa} learns a distortion manifold by aligning representations of similarly degraded images, irrespective of content. Yet, such approaches rigidly collapse embeddings without accounting for perceptual effects like masking \citep{Bovik2013AutomaticPO}, where content alters perceived quality. In practice, perceptual similarity depends jointly on both distortion severity and image content, a dependency that current SSL-IQA models fail to capture in a flexible, scalable manner.

We propose \textbf{SHAMISA} (\textbf{SHA}ped \textbf{M}odeling of \textbf{I}mplicit \textbf{S}tructural \textbf{A}ssociations), a non-contrastive self-supervised framework that addresses this challenge by learning representations jointly sensitive to both distortion and content through explicitly constructed relation graphs. ``SHAped Modeling'' refers to graph-based relational supervision, whereas ``Implicit Structural Associations'' denote the latent perceptual relations encoded in these graphs, providing finer control over similarity learning than prior SSL-IQA methods that treat distorted views uniformly. SHAMISA draws inspiration from ExGRG \citep{exgrg}, which introduced explicit graph-based guidance for self-supervised learning in the graph domain. In contrast, SHAMISA extends this idea to visual data, where quality prediction depends on fine-grained perceptual variations. Prior methods implicitly impose sparse relational structures, resulting in disconnected or inconsistent supervision, which SHAMISA addresses through explicitly shaped relation graphs.

At the core of SHAMISA is a compositional distortion engine that generates infinite compositional degradations from continuous parameter spaces. Each mini-batch is built from reference images that form distortion composition groups where only one degradation factor varies, enabling controlled sampling across content, distortion type, and severity. From this, we build two categories of relation graphs. The \textbf{Metadata-Driven Graphs} encode pairwise similarity from distortion metadata, encouraging images with similar degradations to lie close in the learned manifold while inducing controlled representational shifts as distortion severity varies. The \textbf{Structurally Intrinsic Graphs} are constructed from the latent feature space using $k$-nearest-neighbor (kNN) relations and deep clustering.
These graphs supervise a non-contrastive VICReg-style objective with graph-weighted invariance. With stop-gradient applied to graph construction, each iteration builds relation graphs from the current representations and updates the model under the resulting objective in a single optimization step.

SHAMISA replaces rigid positive-pair rules with a single relational framework for content-dependent and distortion-dependent learning. For downstream evaluation, the learned encoder is frozen and paired with a linear regressor. Its self-supervised pre-training uses neither IQA opinion scores nor a contrastive objective.

\vspace{1ex}
\noindent
Our main contributions are:
\begin{enumerate}
    \item We introduce \textbf{SHAMISA}, a non-contrastive self-supervised framework for NR-IQA that encodes both distortion-aware and content-aware information into a shared representation space via explicit relational supervision.
    \item We propose a distortion engine that generates compositional groups with controlled variation, enabling fine-grained similarity learning across distortion factors and content.
    \item We develop a dual-source relation graph construction strategy that combines Metadata-Driven Graphs based on known degradation metadata with Structurally Intrinsic Graphs derived from the evolving feature space. A stop-gradient alternating update couples on-the-fly graph construction with representation learning.
    \item SHAMISA generalizes the rigid pairing schemes used in prior SSL-IQA methods within a unified graph-weighted invariance framework and achieves strong overall performance across synthetic, authentic, and cross-dataset benchmarks. The results include the strongest overall six-dataset average among the compared SSL methods, improved transfer in the reported cross-dataset comparisons, and favorable qualitative diagnostics.
\end{enumerate}

\section{Related Work}
\label{sec:related_work}
\subsection{Traditional and Supervised NR-IQA}
No-Reference Image Quality Assessment (NR-IQA) aims to estimate perceptual image quality without access to pristine references. Traditional methods rely on handcrafted features derived from Natural Scene Statistics (NSS), modeling distortions as deviations from expected regularities. Examples include BRISQUE \citep{mittal2012no}, NIQE \citep{mittal2013making}, DIIVINE \citep{moorthy2011blind}, and BLIINDS \citep{saad2012blind}, while CORNIA \citep{ye2012unsupervised} and HOSA \citep{xu2016blind} construct codebooks from local patches. Though effective on synthetic distortions, these models often fail on authentically distorted images due to a lack of semantic awareness.

Supervised NR-IQA models typically use pre-trained CNNs (e.g., ResNet \citep{he2016deep}) to extract deep features, which are mapped to quality scores via regression. Techniques like HyperIQA \citep{su2020blindly}, RAPIQUE \citep{tu2021rapique}, and MUSIQ \citep{DBLP:journals/corr/abs-2108-05997} adapt neural architectures for perceptual modeling. PaQ-2-PiQ \citep{ying2020patches} incorporates both image- and patch-level quality labels. Others like DB-CNN \citep{zhang2018blind} and PQR \citep{zeng2017probabilistic} combine multiple feature streams. However, such methods require large-scale and costly human annotations, limiting scalability and generalization.

\subsection{Self-Supervised Learning for NR-IQA}
Self-supervised learning (SSL) enables representation learning from unlabeled data. Contrastive methods like SimCLR \citep{chen2020simple} and MoCo \citep{he2020momentum} rely on sampled negatives, which can introduce false negatives and sampling bias when semantically related samples are treated as negatives \citep{bielak2021graph,chuang2020debiased,chen2022incremental}. Non-contrastive methods like VICReg \citep{vicreg} avoid negatives through invariance and decorrelation. In graph domains, ExGRG \citep{exgrg} introduces explicit relation graphs for SSL, combining domain priors and feature structure. SHAMISA adopts a similar inductive bias but applies it to NR-IQA, where perceptual similarity depends jointly on content and degradation.

Recent SSL-IQA works adapt these paradigms to quality prediction. CONTRIQUE \citep{madhusudana2022image} clusters samples with similar distortion types into classes, learning quality-aware embeddings. QPT \citep{zhao2023quality} aligns patches under shared distortion assumptions. Re-IQA \citep{reiqa} uses two encoders: one pre-trained for content-specific features and another trained for distortion-specific features. Their outputs are concatenated and fed to a single linear regressor. This assumes that a shallow fusion layer can recover complex content-distortion interactions from separately learned streams. Re-IQA may therefore under-represent content-distortion interactions, even though these interactions play a central role in perceptual quality assessment rather than being evaluated in isolation \citep{mohammadi2014subjective, chandler2013seven}.

ARNIQA \citep{arniqa} instead aligns representations of images degraded equally, disregarding content. While it captures distortion similarity, it rigidly enforces uniform proximity and ignores perceptual effects introduced by content. In contrast, SHAMISA learns a unified representation space that respects both distortion and semantic structure through two complementary relation graph families. Metadata-driven and structurally intrinsic graphs together encode distortion-aware similarity and perceptual affinities across content. This provides soft, fine-grained constraints that generalize prior methods as special cases.

Recent opinion-unaware methods expand the forms of supervision available for quality representation learning. GRepQ \citep{grepq} combines low-level and vision-language features; SFD-IQA \citep{sfd_iqa} and HiRQA \citep{hirqa} introduce semantic discrimination or image--text alignment; and TRIQA \citep{triqa} learns from ordered distortion triplets. In the multimodal regime, Refine-IQA \citep{refine_iqa} uses reinforcement fine-tuning for image-quality perception, VITAL \citep{vital} develops image-and-video quality large multimodal models, and W2S-VQA \citep{w2s_vqa} applies weak-to-strong supervision to video quality. These approaches use language alignment, ordered triplets, or multimodal supervision. By contrast, SHAMISA learns a compact visual encoder from unlabeled distorted images through continuous single-factor trajectories and explicit relational supervision, without language alignment or discrete triplet labels.

\subsection{Compositional Distortion Modeling and Relational Supervision}
Degradation engines are central to SSL-IQA training. Prior models apply a limited set of discrete distortions or fixed composition rules \citep{reiqa, zhao2023quality}. ARNIQA, for example, applies sequential distortions with fixed sampling schemes but lacks structured sampling for controlled variation. RealESRGAN \citep{wang2021real} follows a fixed distortion sequence from predefined groups. Such engines offer limited degradation diversity and do not support relational supervision.

SHAMISA introduces a compositional distortion engine that generates continuously parameterized degradations with uncountable variation. We partition each mini-batch into \emph{tiny-batches} (small, fixed-size subsets). Each tiny-batch contains distortion composition groups in which only one degradation factor varies, enabling precise control over severity and type.
This supports precise metadata generation for graph construction. Unlike prior works that enforce binary or fixed similarity constraints, SHAMISA enables structured shifts in representation space by applying soft supervision from graph relations. Similar samples lie closer, while severity variation introduces gradual transitions.

Combined with its dual-source relational supervision, SHAMISA learns distortion-aware and content-sensitive embeddings in a non-contrastive setting. By learning beyond rigid class labels and negative sampling, SHAMISA provides a scalable framework for NR-IQA without using IQA opinion scores during self-supervised pre-training.

\section{Method}
\label{sec:method}
\begin{figure*}[t]
    \centering
    \includegraphics[width=\textwidth]{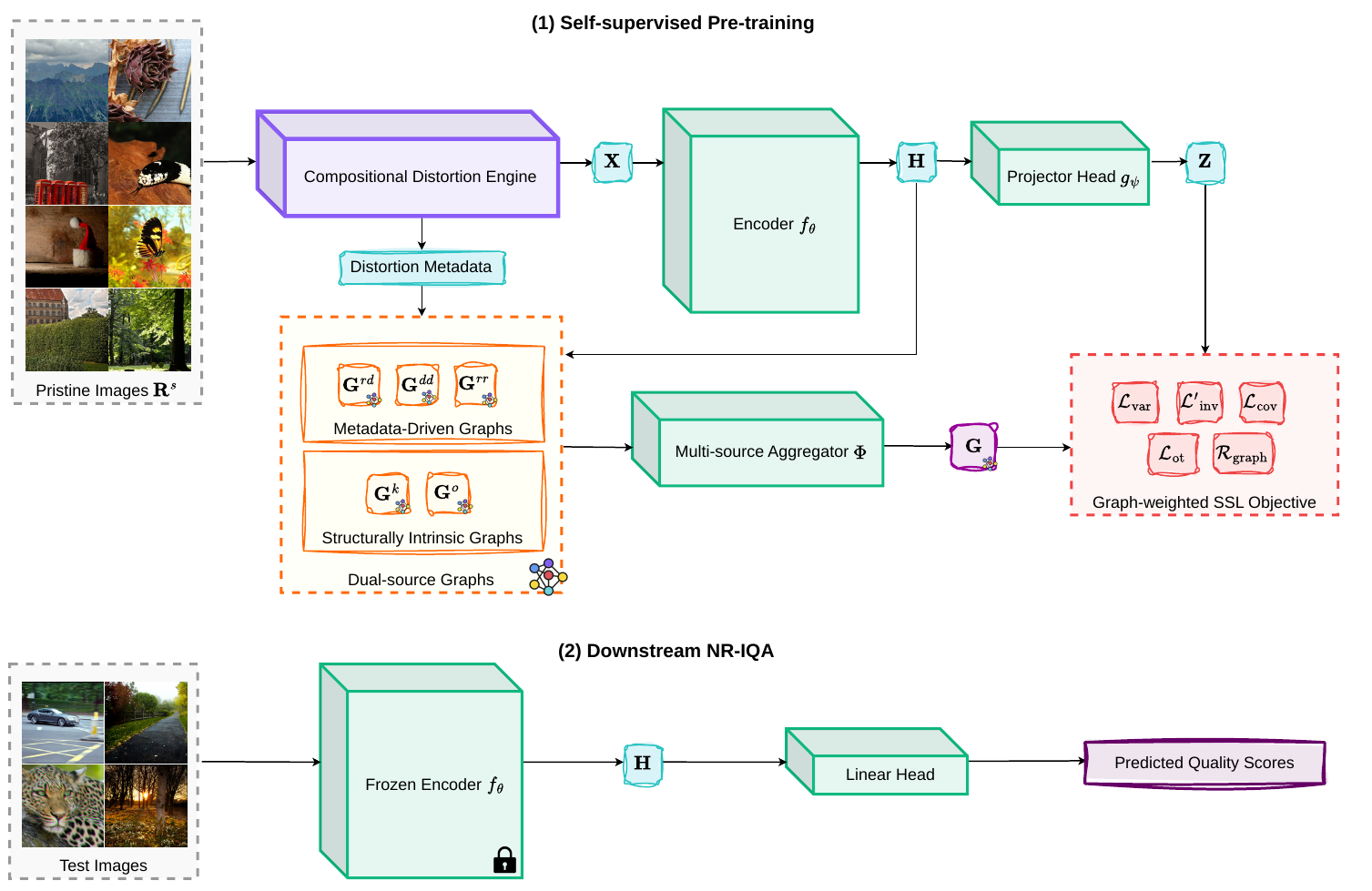}
    \caption{
        \textbf{Overview of the proposed SHAMISA framework.}
        \textbf{(1) Self-supervised pre-training.} Pristine images are transformed by a compositional distortion engine to form a mini-batch $\mathbf{X}$ and distortion metadata. The encoder $f_{\theta}$ and projector $g_\psi$ produce representations $\mathbf{H}$ and embeddings $\mathbf{Z}$. Metadata-driven and structurally intrinsic graphs are aggregated by $\Phi$ into $\mathbf{G}$ to guide the SSL objective.
        \textbf{(2) Downstream NR-IQA.} We freeze $f_{\theta}$ and train a lightweight regressor on $\mathbf{H}$ to predict quality scores.
    }
    \label{fig:pipeline}
\end{figure*}
As illustrated in Fig.~\ref{fig:pipeline}(1), SHAMISA combines structured distortion generation with dual-source relational supervision to learn quality-aware features in a non-contrastive setting. During downstream evaluation in Fig.~\ref{fig:pipeline}(2), we transfer the learned encoder to NR-IQA by freezing backbone representations and fitting a lightweight regressor, so downstream performance reflects representation quality rather than head complexity.

\subsection{Overview and SSL Formulation for NR-IQA}
SHAMISA comprises three core components: controlled single-factor distortion trajectories, metadata-driven and structurally intrinsic relation graphs, and graph-weighted VICReg. We instantiate these components by pre-training a ResNet-50 encoder $f_\theta$ with a 2-layer MLP projector $g_\psi$ on images degraded online by the distortion engine. The VICReg-style non-contrastive objective \citep{vicreg} avoids negative sampling and its associated sampling bias \citep{bielak2021graph}. After self-supervised pre-training, we discard $g_\psi$, freeze $f_\theta$ without fine-tuning, and train a linear regressor on the frozen features for quality prediction.

\noindent\emph{Tensorized batch construction.}
Within each mini-batch, let $\mR^s$ stack $N_{\text{ref}}$ pristine crops. For compact notation, $\mX^{(d)}=\chi_d(\mR^s)$ collects the degraded crops associated with composition index $d\in\{1,\dots,N_{\text{comp}}\}$, with each tiny-batch using its own sampled compositions (Sec.~\ref{sec:distortion_engine}). Stacking the pristine and degraded views gives
\begin{equation}
    \begin{aligned}
        \mX & = [\,\mR^s,\ \mX^{(1)},\ \dots,\ \mX^{(N_{\text{comp}})}\,]
        \in \R^{N \times 3 \times H_{in} \times W_{in}},                  \\
        N   & = N_{\text{ref}}\,(N_{\text{comp}}+1).
    \end{aligned}
    \label{eq:batch_concat}
\end{equation}
Representations $\mH$ and embeddings $\mZ$ are
\[
    \mH = f_\theta(\mX) \in \R^{N \times d_h},\qquad
    \mZ = g_\psi(\mH) \in \R^{N \times d_z}.
\]
Human opinion scores are used only for the regression head after pre-training. The exact crop construction and tensor shapes are given in Supplementary Appendix A.

\subsection{Compositional Distortion Engine}
\label{sec:distortion_engine}
A \emph{distortion function} is one atomic degradation from KADID-10K with 24 functions across 7 categories (Brightness change, Blur, Spatial, Noise, Color, Compression, Sharpness \& Contrast) \citep{kadid10k}. A \emph{distortion composition} in SHAMISA is an \emph{ordered composition} of multiple distortion functions with at most one function per category, applied sequentially. Unlike prior SSL-IQA setups based on discrete severity grids \citep{zhao2023quality, arniqa}, SHAMISA samples \emph{continuous} severities, yielding an \emph{uncountable} family of compositions. Order is perceptually consequential, hence we randomize it. In practice each iteration uses a finite $N_{\text{comp}}$, but randomization across iterations explores novel compositions, and continuous level differences are exploited later by our relation graphs.

\paragraph{Severity calibration}
For each distortion function $f_m$ we map its native parameter domain $[\lambda_m^{\min},\,\lambda_m^{\max}]$ into $[0,1]$ via a piecewise-linear calibration $\rho_m:[\lambda_m^{\min},\,\lambda_m^{\max}]\to[0,1]$ obtained by linearly interpolating the discrete intensities used in \citep{kadid10k}. This yields comparable normalized severities across functions. Sampling operates in the normalized space, and native parameters are recovered through $\rho_m^{-1}$ when required.

\paragraph{Sampling a composition}
For each distortion composition we sample a tuple $(M,\,S,\,f,\,\pi,\,\lambda)$. The variables are defined below.
$M\sim\mathrm{Unif}\{1,\dots,M_d\}$ is the number of distortion \emph{functions} in the composition with at most one per category.
$S\subset\{1,\dots,7\}$ is a size-$M$ subset indexing the seven KADID-10K distortion categories, sampled uniformly without replacement.
$f=(f_1,\dots,f_M)$ selects one function per category in $S$.
$\pi$ is a uniformly sampled permutation of $\{1,\dots,M\}$ that specifies the application order.
The vector $\lambda=(\lambda_1,\dots,\lambda_M)\in[0,1]^M$ contains per-function normalized severities sampled i.i.d. by drawing $\epsilon_m\sim\mathcal{N}(0,0.5^2)$ and setting $\lambda_m=\min(1,|\epsilon_m|)$, where $[0,1]^M$ is the $M$-dimensional unit hypercube.
Supplementary Appendix B gives the resulting distribution and its uniform-sampling ablation.
Given an input image $x$, let $\theta_m \coloneqq \rho_m^{-1}(\lambda_m)$ and write $f_m(\,\cdot\mid\theta_m)$.
The composed degradation is
\begin{equation}
    \mathcal{C}(x\mid\pi,f,\lambda)
    \;=\;
    \big(f_{\pi(M)}(\,\cdot\mid\theta_{\pi(M)}) \circ \cdots \circ f_{\pi(1)}(\,\cdot\mid\theta_{\pi(1)})\big)(x).
    \label{eq:composition}
\end{equation}

\subsubsection{Single-factor variation and trajectories}
Each mini-batch is partitioned into $B$ tiny-batches. We use $[n]\coloneqq\{1,\dots,n\}$ for index ranges, so $i\in[B]$ indexes tiny-batches. In tiny-batch $i$, select $R$ references $\{r_{i,j}\}_{j=1}^{R}$ and instantiate $C$ \emph{distortion composition groups}. For the $k$-th \textit{composition group} with $k\in[C]$, sample a base \textit{composition} $(M_{i,k},S_{i,k},f_{i,k},\pi_{i,k},\lambda_{i,k}^{(0)})$, where $\lambda_{i,k}^{(0)}\in[0,1]^{M_{i,k}}$ is the base severity vector. Choose exactly one distortion function whose level will vary, indexed by $m^\star(i,k)\in\{1,\dots,M_{i,k}\}$, and generate $L$ severity levels $\{s_{i,k}^{(l)}\}_{l=1}^{L}\subset[0,1]$. We draw each level i.i.d. as $s_{i,k}^{(l)}=\min(1,|\epsilon^{(l)}|)$ with $\epsilon^{(l)}\sim\mathcal{N}(0,0.5^2)$.
Define the per-level severities component-wise by
\[
    (\lambda_{i,k}^{(l)})_m \;=\;
    \begin{cases}
        (\lambda_{i,k}^{(0)})_m, & m \neq m^\star(i,k), \\[2pt]
        s_{i,k}^{(l)},           & m = m^\star(i,k).
    \end{cases}
\]
This single-factor scheme varies one distortion function while holding the others fixed, producing controlled severity variation that isolates that function’s effect in the learned representation. Formally, applying the level-$l$ composition in group $k$ to reference $r_{i,j}$ yields the distorted image

\begin{equation}
    \begin{aligned}
        x_{i,j,k,l} & = \mathcal{C}\big(r_{i,j}\mid\pi_{i,k}, f_{i,k}, \lambda_{i,k}^{(l)}\big), \\
                    & \qquad i\!\in\![B],\ j\!\in\![R],\ k\!\in\![C],\ l\!\in\![L].
    \end{aligned}
    \label{eq:engine_gen}
\end{equation}
Consequently, the mini-batch contains $N_{\text{ref}}=BR$ references and each reference receives $N_{\text{comp}}=CL$ compositions, with sampling performed independently for each tiny-batch.

\subsection{Explicit Relation Graphs and Graph-weighted VICReg}
We adopt the variance and covariance regularizers of VICReg \citep{vicreg} and replace its augmentation-paired invariance term $\mathcal{L}_{\mathrm{inv}}$ with a graph-weighted variant $\mathcal{L}'_{\mathrm{inv}}$, in the spirit of explicitly generated relation graphs \citep{exgrg}:
\begin{align}
    \mathcal{L}_{\mathrm{var}}  & = \sum_{t=1}^{d_z} \max \big(0,\, 1 - \sqrt{\Cov(\mZ)_{t,t}}\big), \label{eq:var}                \\
    \mathcal{L}_{\mathrm{cov}}  & = \sum_{t=1}^{d_z} \sum_{\substack{u=1                                                           \\ u\neq t}}^{d_z} \Cov(\mZ)^2_{t,u}, \label{eq:cov}\\
    \mathcal{L}_{\mathrm{inv}}  & = \sum_{(i,j)\in\mathcal{A}} \big\|\mZ_{i,.}-\mZ_{j,.}\big\|_2^2, \label{eq:aug_inv}             \\
    \mathcal{L}'_{\mathrm{inv}} & = \sum_{i=1}^{N}\sum_{j=1}^{N} \mG_{i,j}\, \big\|\mZ_{i,.}-\mZ_{j,.}\big\|_2^2. \label{eq:g_inv}
\end{align}
Here $\mathcal{A}\subseteq [N]\times[N]$ denotes the set of augmentation positives, and $\mG\in[0,1]^{N\times N}$ is the weighted adjacency matrix of a soft relation graph (Sec.~\ref{sec:aggregation}), where each entry specifies how strongly the corresponding pair is encouraged to remain invariant.
These soft, controllable entries of $\mG$ instantiate the implicit structural associations, encoding relational cues that jointly capture content and distortion. Intuitively, $\mathcal{L}_{\mathrm{var}}$ enforces a minimum per-dimension variance to avoid collapse, $\mathcal{L}_{\mathrm{cov}}$ decorrelates dimensions, and $\mathcal{L}'_{\mathrm{inv}}$ brings together embeddings proportionally to relational strengths encoded by $\mG$. In contrast, $\mathcal{L}_{\mathrm{inv}}$ uses a \emph{rigid} binary set $\mathcal{A}$ of augmentation positives. When $\mG$ is the binary indicator of $\mathcal{A}$ (i.e., $\mG_{i,j}=1$ iff $(i,j)\in\mathcal{A}$ and $0$ otherwise), Eq.~\ref{eq:g_inv} reduces to VICReg’s original invariance in Eq.~\ref{eq:aug_inv}.

Our self-supervised pre-training objective is
\begin{equation}
    \mathcal{L}_{\mathrm{ssl}} \;=\; \alpha\,\mathcal{L}_{\mathrm{var}} + \beta\,\mathcal{L}_{\mathrm{cov}} + \gamma\,\mathcal{L}'_{\mathrm{inv}}.
    \label{eq:pretext}
\end{equation}
SHAMISA remains \emph{non-contrastive}: we do not use negatives and instead prevent collapse via the variance-covariance terms \citep{vicreg}, rather than through negative pairs as in contrastive methods \citep{chen2020simple}.

\paragraph{Relation to prior positives}
Cosine invariance underlies InfoNCE in SimCLR \citep{chen2020simple}. With $\ell_2$-normalized features, $\|a-b\|_2^2 = 2\,(1-\cos(a,b))$, so Euclidean and cosine invariances are equivalent up to scaling.
Thus, prior invariances that rely on \emph{rigid binary pairings} become special cases of SHAMISA’s \emph{soft} graph-weighted invariance. ARNIQA pairs images with the same distortion type and level across different contents \citep{arniqa}. Re-IQA pairs overlapping crops of the same image \citep{reiqa}. CONTRIQUE uses synthetic distortion classes or instance discrimination for UGC \citep{madhusudana2022image, zhao2023quality}. Each choice is recovered by setting $\mG$ to the corresponding sparse binary adjacency.

    {\makeatletter\def\@IEEEsectpunct{\ }\paragraph{Why an explicit \texorpdfstring{$\mG$}{G}?}\makeatother
        \emph{Laplacian Eigenmap view.} A sparse augmentation-positive graph can split into disconnected components and yield a rank-deficient Laplacian \citep{balestriero2022contrastive}. Soft cross-component edges in $\mG$ alleviate this limitation and provide a better posed invariance term \citep{exgrg}.
        \emph{Alternating update view.} We construct $\mG$ from the current representations to define soft relational targets, then update $f_\theta$ and $g_\psi$. A stop-gradient on the graph inputs, defined formally in Sec.~\ref{sec:aggregation}, allows graph construction and parameter updates within one optimization step \citep{chen2021exploring, exgrg}.
    }

\subsection{Metadata-Driven Graphs}\label{sec:metadata_graphs}
We convert engine metadata into soft relational weights that shape the embedding space through two complementary effects. Within a given content, similarity to the pristine anchor decreases smoothly as severity increases. Across contents, samples from the same distortion composition group remain neighbors when their severities are close.

We next define two monotone maps $\phi,\hat\phi:[0,1]\to[0,1]$, with $\phi'(u)\le 0$ and $\hat\phi'(u)\le 0$, that convert a normalized severity or severity gap into a similarity weight. We set $\phi(u)=\hat\phi(u)=\exp(-\kappa u)$ with $\kappa=3$. Both maps are bounded in $[0,1]$ and strictly decrease with $u$.

\subsubsection{Reference-distorted graph \texorpdfstring{$\mG^{rd}$}{Grd}}
Recall that in tiny-batch \(i\) and group \(k\), \(m^\star(i,k)\) is the varying function and \(\lambda^{(l)}_{i,k}\in[0,1]^{M_{i,k}}\) is the level-\(l\) severity vector. We assign an edge from each reference to \emph{each} of its degraded versions:
\begin{equation}
    \mG^{rd}\big[r_{i,j},\, x_{i,j,k,l}\big] \;=\; \phi\Big((\lambda^{(l)}_{i,k})_{m^\star(i,k)}\Big).
    \label{eq:grd}
\end{equation}
\emph{Intuition.} Small severities yield $\phi\!\approx\!1$ and keep mild degradations near the pristine anchor. As severity increases, $\phi$ decreases smoothly to its minimum at the maximum level, weakening the attraction and allowing stronger degradations to move farther away. This gives a graded severity-aware relation instead of a rigid binary pair.

\subsubsection{Distorted-distorted graph \texorpdfstring{\(\mG^{dd}\)}{Gdd}}
Within the same distortion composition group \((i,k)\), define the 1D severity gap
\[
    \delta_{i,k}(l,l') \;=\; \big|\, s^{(l)}_{i,k} - s^{(l')}_{i,k} \,\big| \in [0,1],
\]
and, for any \(j,j'\in[R]\), define the edge weight between two distorted images whose contents may differ as
\begin{equation}
    \mG^{dd} \big[x_{i,j,k,l},\, x_{i,j',k,l'}\big] \;=\; \hat\phi \Big(\delta_{i,k}(l,l')\Big).
    \label{eq:gdd}
\end{equation}
\emph{Intuition.} Nearby severity levels attract, while distant levels do not. This ties together similarly degraded images \emph{regardless of content}, complementing \(\mG^{rd}\). To avoid density, we apply a \emph{global} Top-\(K_d\) sparsifier.
\[
    \mG^{dd} \;\leftarrow\; F_{K_d}(\mG^{dd}),
\]
where \(F_{K_d}\) retains the \(K_d\) largest entries \emph{in the whole matrix} and zeros the rest.

\subsubsection{Reference-reference graph \texorpdfstring{\(\mG^{rr}\)}{Grr}}
We weakly connect pristine images across contents to stabilize a common high-quality anchor:
\begin{equation}
    \mG^{rr}\big[r_{i,j},\, r_{i',j'}\big] \;=\; w_{rr},\qquad w_{rr}\in(0,1].
    \label{eq:grr}
\end{equation}
\emph{Intuition.} This graph builds a coherent pristine neighborhood across contents without forcing content collapse. The variance and covariance regularizers prevent trivial solutions while these links stabilize a shared high-quality anchor in the representation space.

The three metadata graphs provide \emph{severity-aware}, \emph{content-aware} soft relations that are fed to the overall relation aggregation (Sec.~\ref{sec:aggregation}) and used inside the graph-weighted invariance term (Eq.~\ref{eq:g_inv}) to shape quality-sensitive embeddings.

\subsection{Structurally Intrinsic Graphs}\label{sec:structure_graphs}
Beyond metadata, we inject \emph{emergent} structure from the evolving representation space. These graphs encourage local manifold smoothness and global prototype organization, complementing the metadata graphs.

\subsubsection{kNN relation graph}
Compute similarities on encoder features $\mH$ using the reciprocal Euclidean affinity of their $\ell_2$-normalized representations:
\[
    \bar{\mH}_{i,.}=\frac{\mH_{i,.}}{\|\mH_{i,.}\|_2},
    \qquad
    \mS^{H}_{i,j}=\frac{1}{1+\|\bar{\mH}_{i,.}-\bar{\mH}_{j,.}\|_2}.
\]
Form the kNN graph by keeping, for each sample, the top \(k_n\) similarity scores and removing self-connections:
\begin{equation}
    \mG^{k} \,=\, F_{k_{n}}(\mS^{H}) \in \R^{N\times N},
    \label{eq:gknn}
\end{equation}
where \(F_{k_{n}}\) retains the largest \(k_{n}\) entries in each row of \(\mS^{H}\) and zeros the rest.

\emph{Intuition.} Deep features are well correlated with perceptual similarity \citep{zhang2018unreasonable}. Thus, nearest neighbors in \(\mH\) define perceptually coherent local neighborhoods. \(\mG^{k}\) promotes local smoothness of the embedding with respect to this current geometry.

\subsubsection{OT-based clustering guidance}
Following ExGRG \citep{exgrg}, we adapt optimal-transport (OT) clustering guidance to IQA. Let \(\mC\in\R^{K\times d_h}\) be $K$ trainable prototypes and \(\mA\in\R^{N\times K}\) the soft \emph{assignments} computed from the row-normalized features $\bar{\mH}$ and prototypes $\bar{\mC}$:
\begin{equation}
    \mA_{j,k} \,=\, \frac{\exp(\bar{\mH}_{j,.}^\top \bar{\mC}_{k,.}/\tau_c)}{\sum_{k'} \exp(\bar{\mH}_{j,.}^\top \bar{\mC}_{k',.}/\tau_c)}.
    \label{eq:assign}
\end{equation}
Balanced \emph{targets} \(\mT\in\R^{N\times K}\) are obtained by a Sinkhorn-Knopp solver with entropy regularization \citep{cuturi2013sinkhorn}, which enforces approximately uniform cluster marginals per mini-batch \citep{asano2019self, caron2020unsupervised}. For probability vectors \(\mathbf{t}, \mathbf{a} \in [0,1]^K\) with \(\sum_{k=1}^K t_k = \sum_{k=1}^K a_k = 1\), we define the cross-entropy
\[
    \mathrm{CE}(\mathbf{t},\mathbf{a}) \;\coloneqq\; -\sum_{k=1}^{K} t_k \log a_k.
\]
For notational convenience, we write \(\mT_{x_{i,j,k,l}}\) and \(\mA_{x_{i,j,k,l}}\) for the rows of \(\mT\) and \(\mA\) associated with distorted image \(x_{i,j,k,l}\) in Eq.~\ref{eq:engine_gen}.
The OT loss has two contributions,
\begin{equation}
    \mathcal{L}_{\mathrm{ot}}
    =
    \frac{1}{N}\sum_{n=1}^{N}
    \mathrm{CE}\big(\mT_{n,.},\,\mA_{n,.}\big)
    +
    \frac{1}{|\mathcal{P}|}
    \sum_{(u,v)\in\mathcal{P}}
    \mathrm{CE}\big(\mT_{u,.},\,\mA_{v,.}\big),
    \label{eq:otloss}
\end{equation}
\[
    \mathcal{P}
    \;=\;
    \big\{(x_{i,j,k,l},x_{i,j',k,l}) : i\in[B],\, k\in[C],\, l\in[L],\, j\neq j' \big\}.
\]
where $N$ is the mini-batch size in Eq.~\ref{eq:batch_concat}. The first term aligns each sample’s assignment to its own balanced target over all $N$ images in the mini-batch. The set \(\mathcal{P}\) contains cross-content distorted-image pairs that share the same tiny-batch \(i\), composition group \(k\), and severity level \(l\). The second term averages the corresponding cross-entropy over these pairs.

Soft assignments \(\mA\) provide graded membership probabilities, which reduce premature hard assignments and allow clusters to form more flexibly.
From \(\mA\) we compute an assignment-affinity matrix
\[
    \mS^{A}_{i,j} \,=\, \sum_{k} \mA_{i,k}\,\log \mA_{j,k},
\]
then normalize entries and apply a global sparsity limit to obtain the relation graph
\begin{equation}
    \mG^{o} \,=\, F_{K_g}\big(\mathrm{norm}(\mS^{A})\big) \in \R^{N\times N},
    \label{eq:go}
\end{equation}
where \(\mathrm{norm}\) maps the matrix entries into \([0,1]\) via global min-max normalization, and \(F_{K_g}\) keeps the globally largest \(K_g\) entries, zeroing the rest.

\emph{Intuition.} OT-based guidance shapes the global topology by encouraging samples with similar prototype memberships to remain close across contents, while soft assignments preserve uncertainty during cluster formation. Prototypes \(\mC\) and the encoder \(f_\theta\), and hence \(\mH\), are updated through \(\mathcal{L}_{\mathrm{ot}}\) together with the overall objective. The assignments \(\mA\) and targets \(\mT\) are recomputed from the current features at each iteration.

\subsection{Multi-Source Aggregation with Stop-Gradient}\label{sec:aggregation}
Let
\[
    S_G \,=\, \{\mG^{rd},\, \mG^{dd},\, \mG^{rr},\, \mG^{k},\, \mG^{o}\}.
\]
We apply a stop-gradient to each source and aggregate the resulting graphs with \(\Phi\):
\begin{align}
    &\tilde{\mG}^{(i)} = \mathrm{sg}\big(\mG^{(i)}\big),                       \\
    &\mG = \sum_{\mG^{(i)}\in S_G} \omega_i\, \tilde{\mG}^{(i)},
    \qquad \omega_i \ge 0,\ \sum_i \omega_i = 1,
    \label{eq:gagg}
\end{align}
where $\mathrm{sg}(\cdot)$ is the stop-gradient operator (identity in forward, zero gradient in backward). The weights \(\omega\) are produced by a lightweight \emph{hypernetwork} \(\Phi\) \citep{ha2016hypernetworks, exgrg} from simple per-graph statistics,
\begin{equation}
    \begin{aligned}
        \omega      & = \mathrm{softmax}\Big(\Phi\big(\{\zeta^{(i)}\}_{i=1}^{T}\big)\Big), \\
        \zeta^{(i)} & = \big(\mathrm{sum}(\tilde{\mG}^{(i)}),\ \mathrm{nnz}(\tilde{\mG}^{(i)})\big),
    \end{aligned}
    \label{eq:omegas}
\end{equation}
where \(T=|S_G|\), $\{\zeta^{(i)}\}_{i=1}^{T}$ denotes stacking the per-graph statistics over sources, \(\mathrm{sum}(\cdot)\) is the total edge weight (connection mass), and \(\mathrm{nnz}(\cdot)\) is the number of nonzero entries (sparsity). This hypernetwork aggregates an arbitrary set of sources online, adapting \(\omega\) as the graphs evolve during training. The two statistics are inexpensive, scale-agnostic proxies for \emph{strength} and \emph{coverage} of each source, providing a practical signal for allocating weight without hand-tuning.

\subsection{Training Details}
We maintain sparse relation graphs via the pruning operators $F_{k_n}$ for row-wise Top-$k$ and $F_{K_d}, F_{K_g}$ for global Top-$K$. The graph-weighted invariance $\mathcal{L}'_{\mathrm{inv}}$ is evaluated only over nonzero edges rather than all $N^2$ pairs. This respects sparsification and reduces computation.
We include the graph regularizer
\[
    \mathcal{R}_{\mathrm{graph}} \;=\; -\sum_{i,j} \mG_{i,j}^{\,2},
\]
to encourage nontrivial mass on the retained edges after sparsification. Under the simplex constraint on $\omega$, this favors informative retained edge weights instead of vanishing graph mass. Our construction choices keep all graph entries in $[0,1]$, which avoids uncontrolled saturation.

\subsection{End-to-End Optimization and Regression Protocol}
The full objective is
\begin{equation}
    \mathcal{L}_{\mathrm{total}} \,=\, \mathcal{L}_{\mathrm{ssl}}
    \;+\; \eta\,\mathcal{L}_{\mathrm{ot}} \;+\; \xi\,\mathcal{R}_{\mathrm{graph}}.
    \label{eq:total}
\end{equation}
Algorithm~\ref{alg:shamisa_pretraining} summarizes the repeated pre-training loop. At each update, it constructs $\mG$ from the current representations and then updates the encoder $f_\theta$, projector $g_\psi$, prototype parameters, and aggregation hypernetwork $\Phi$ by minimizing Eq.~\ref{eq:total}, with stop-gradient applied to the graph inputs as defined in Sec.~\ref{sec:aggregation}.

\begin{algorithm}[t]
    \caption{SHAMISA self-supervised pre-training}
    \label{alg:shamisa_pretraining}
    \begin{algorithmic}[1]
        \REQUIRE Pristine training images; initialized $f_\theta$, $g_\psi$, $\Phi$, and prototypes
        \FOR{each mini-batch}
            \STATE Sample pristine references and distortion compositions; generate $\mX$ using Eqs.~\ref{eq:composition} and~\ref{eq:engine_gen}.
            \STATE Compute $\mH\gets f_\theta(\mX)$ and $\mZ\gets g_\psi(\mH)$.
            \STATE Construct metadata-driven graphs $\{\mG^{rd},\mG^{dd},\mG^{rr}\}$ and structurally intrinsic graphs $\{\mG^{k},\mG^{o}\}$.
            \STATE Aggregate the relation graphs with $\Phi$ using Eqs.~\ref{eq:gagg} and~\ref{eq:omegas} to obtain $\mG$.
            \STATE Evaluate $\mathcal{L}_{\mathrm{total}}$ in Eq.~\ref{eq:total} and update the trainable parameters, stopping gradients at the graph inputs.
        \ENDFOR
        \STATE Discard $g_\psi$ and freeze $f_\theta$ for downstream linear probing.
    \end{algorithmic}
\end{algorithm}

The exact crop and tensor construction, selected dimensions, and optimization settings are implementation details consolidated in Supplementary Appendix A. Top-$K$ pruning and sparse edge evaluation reduce computational cost while preserving the relational objective and avoiding dense $N^2$ evaluation. These choices complete the implementation without changing the conceptual components described above.

\section{Experimental Evaluation}
\label{sec:experiments}
We evaluate SHAMISA on standard NR-IQA benchmarks with a linear-probe protocol \citep{madhusudana2022image,reiqa,arniqa}, report SRCC and PLCC, and study within-dataset performance, cross-dataset transfer, gMAD diagnostics, representation visualizations, ablations, and a full-reference extension.

\subsection{Setup}
\label{subsec:setup}
\paragraph{Backbone and pretext}
We initialize a ResNet-50 encoder \(f_\theta\) from ImageNet and \emph{pre-train} it self-supervised with \(\mathcal{L}_{\mathrm{total}}\) in Eq.~\ref{eq:total}. A 2-layer MLP projector \(g_\psi\) is used only during pre-training. Unlabeled inputs are degraded online by the compositional distortion engine in Sec.~\ref{sec:method}. After pre-training, \(g_\psi\) is discarded and \(f_\theta\) remains frozen during downstream evaluation.

\paragraph{Training details}
We pre-train on the 140k KADIS reference images with SGD. Encoder and projector dimensions are \(d_h{=}2048\) and \(d_z{=}256\). Each update uses \(B{=}2\) tiny-batches, \(R{=}3\) references, \(C{=}4\) composition groups, and \(L{=}5\) severity levels. Hence, \(N_{\text{ref}}{=}BR{=}6\), \(N_{\text{comp}}{=}CL{=}20\), and \(N{=}N_{\text{ref}}(N_{\text{comp}}+1){=}126\) in Eq.~\ref{eq:batch_concat}.
Distortion generation uses \(M_d{=}7\). Graph sparsity and OT settings are \(k_n{=}31\), \(K_d{=}4096\), \(K{=}32\), and \(\tau_c{=}0.1\). The OT sparsification budget scales with mini-batch size through \(K_g{=}8N\). Crop size is \(224\times224\). Remaining details are listed in \suppAppArchitectures.

\subsection{Datasets and Evaluation Protocol}
\label{subsec:datasets-protocol}
\paragraph{Pre-training data}
Following ARNIQA \citep{arniqa}, we use the 140k pristine images from KADIS \citep{kadid10k} as content sources to enable a fair comparison under a shared pre-training corpus. We generate degradations online with our compositional distortion engine.

\paragraph{Evaluation datasets}
We include both synthetic and in-the-wild benchmarks. The synthetic datasets are LIVE \citep{sheikh2006statistical}, CSIQ \citep{larson2010most}, TID2013 \citep{ponomarenko2013color}, and KADID-10K \citep{kadid10k}. The in-the-wild datasets are
FLIVE \citep{ying2020patches}, SPAQ \citep{fang2020perceptual}. Brief dataset statistics are available in \suppAppDatasets.

\paragraph{Protocol}
Following standard SSL-IQA practice \citep{madhusudana2022image,reiqa,arniqa}, we create 10 training, validation, and test splits for LIVE, CSIQ, TID2013, KADID-10K, and SPAQ. Synthetic datasets use reference-disjoint sampling, while SPAQ uses random 70, 10, and 20 percent partitions. For FLIVE, we use the official split \citep{ying2020patches}. We train an \(\ell_2\) ridge regressor \citep{hoerl1970ridge} on frozen features and select \(\alpha_{\mathrm{ridge}}\) from a log-spaced grid of 100 values in \([10^{-3},10^{3}]\) by maximizing median validation SRCC across the available splits. At each crop position, we concatenate full- and half-scale features \citep{madhusudana2022image}, apply the ridge head, and then average the five scores. The crop positions are the four corners and the center \citep{zhao2023quality}. We report the median over the repeated splits where applicable. PLCC is computed after the standard four-parameter logistic mapping \citep{VQEG2000}.

\subsection{Compared Methods}
\label{subsec:baselines}
We compare against traditional NR-IQA (BRISQUE \citep{mittal2012no}, NIQE \citep{mittal2013making}), codebook models (CORNIA \citep{ye2012unsupervised}, HOSA \citep{xu2016blind}), and deep supervised NR-IQA (DB-CNN \citep{zhang2018blind}, HyperIQA \citep{su2020blindly}, TReS \citep{golestaneh2022no}, Su et al. \citep{su2023distortion}). We also include SSL-IQA baselines CONTRIQUE \citep{madhusudana2022image}, Re-IQA \citep{reiqa}, and ARNIQA \citep{arniqa}, matching the methods reported in Table~\ref{tab:main_results}. The table retains the coherent six-benchmark comparison assembled in prior literature~\citep{madhusudana2022image,reiqa,arniqa}. We reproduce the ARNIQA entries with its official open-source implementation, on which our evaluation implementation is based; the remaining baseline values are those reported in the cited comparison literature. The recent opinion-unaware image models~\citep{grepq,sfd_iqa,triqa,hirqa} and multimodal quality models~\citep{refine_iqa,vital,w2s_vqa} use different backbones, supervision, datasets, or test protocols. We therefore discuss them qualitatively rather than insert their published scores into Table~\ref{tab:main_results} as if they were obtained under a common protocol.

\subsection{Within-Dataset Performance}

\label{subsec:main-results}
\begin{table*}[t]
    \centering
    \caption{\textbf{Performance comparison on synthetic and in-the-wild NR-IQA datasets.} We report SRCC and PLCC. Best and second-best results in each column are \textbf{bold} and \underline{underlined}. The comparison consolidates results reported by CONTRIQUE, Re-IQA, and ARNIQA \citep{madhusudana2022image,reiqa,arniqa}. We reproduce the ARNIQA row using its official implementation. The remaining baseline rows retain the published values in the cited literature. SHAMISA follows the established split, multi-scale feature, crop-pooling, and regression protocol in Sec.~\ref{subsec:datasets-protocol}.}
    \label{tab:main_results}
    \resizebox{\linewidth}{!}{
        \begin{tabular}{lccccccccccccccc}
            \toprule
                                                   &                   & \multicolumn{8}{c}{Synthetic distortions} & \multicolumn{4}{c}{Authentic distortions} & \multicolumn{2}{c}{}                                                                                                                                                                                                                \\
            \cmidrule(lr){3-10} \cmidrule(lr){11-14}
            Method                                 & Type              &
            \multicolumn{2}{c}{LIVE}               &
            \multicolumn{2}{c}{CSIQ}               &
            \multicolumn{2}{c}{TID2013}            &
            \multicolumn{2}{c}{KADID-10K}          &
            \multicolumn{2}{c}{FLIVE}              &
            \multicolumn{2}{c}{SPAQ}               &
            \multicolumn{2}{c}{Average}                                                                                                                                                                                                                                                                                                                                                              \\
                                                   &                   & SRCC                                      & PLCC                                      & SRCC                 & PLCC              & SRCC              & PLCC              & SRCC              & PLCC              & SRCC              & PLCC              & SRCC              & PLCC              & SRCC              & PLCC \\
            \midrule
            BRISQUE~\citep{mittal2012no}           & Handcrafted
                                                   & 0.939             & 0.935                                     & 0.746                                     & 0.829                & 0.604             & 0.694             & 0.528             & 0.567             & 0.288             & 0.373             & 0.809             & 0.817             & 0.652             & 0.703                    \\
            NIQE~\citep{mittal2013making}          & Handcrafted
                                                   & 0.907             & 0.901                                     & 0.627                                     & 0.712                & 0.315             & 0.393             & 0.374             & 0.428             & 0.211             & 0.288             & 0.700             & 0.709             & 0.522             & 0.572                    \\
            \midrule
            CORNIA~\citep{ye2012unsupervised}      & Codebook
                                                   & 0.947             & 0.950                                     & 0.678                                     & 0.776                & 0.678             & 0.768             & 0.516             & 0.558             & --                & --                & 0.709             & 0.725             & --                & --                       \\
            HOSA~\citep{xu2016blind}               & Codebook
                                                   & 0.946             & 0.950                                     & 0.741                                     & 0.823                & 0.735             & 0.815             & 0.618             & 0.653             & --                & --                & 0.846             & 0.852             & --                & --                       \\
            \midrule
            DB-CNN~\citep{zhang2018blind}          & Supervised
                                                   & 0.968             & 0.971                                     & 0.946                                     & 0.959                & 0.816             & 0.865             & 0.851             & 0.856             & 0.554             & 0.652             & 0.911             & 0.915             & 0.841             & 0.870                    \\
            HyperIQA~\citep{su2020blindly}         & Supervised
                                                   & 0.962             & 0.966                                     & 0.923                                     & 0.942                & 0.840             & 0.858             & 0.852             & 0.845             & 0.535             & 0.623             & \underline{0.916} & 0.919             & 0.838             & 0.859                    \\
            TReS~\citep{golestaneh2022no}          & Supervised
                                                   & 0.969             & 0.968                                     & 0.922                                     & 0.942                & 0.863             & 0.883             & 0.859             & 0.858             & 0.554             & 0.625             & --                & --                & --                & --                       \\
            Su et al.~\citep{su2023distortion}     & Supervised
                                                   & \underline{0.973} & \underline{0.974}                         & 0.935                                     & 0.952                & 0.815             & 0.859             & 0.866             & 0.874             & --                & --                & --                & --                & --                & --                       \\
            \midrule
            CONTRIQUE~\citep{madhusudana2022image} & SSL+LR
                                                   & 0.960             & 0.961                                     & 0.942                                     & 0.955                & 0.843             & 0.857             & \textbf{0.934}    & \textbf{0.937}    & 0.580             & 0.641             & 0.914             & 0.919             & 0.862             & 0.878                    \\
            Re-IQA~\citep{reiqa}                   & SSL+LR
                                                   & 0.970             & 0.971                                     & 0.947                                     & 0.960                & 0.804             & 0.861             & 0.872             & 0.885             & \textbf{0.645}    & \textbf{0.733}    & \textbf{0.918}    & \textbf{0.925}    & 0.859             & 0.889                    \\
            ARNIQA~\citep{arniqa}                  & SSL+LR
                                                   & 0.966             & 0.970                                     & \underline{0.962}                         & \underline{0.973}    & \underline{0.880} & \underline{0.901} & 0.908             & 0.912             & 0.595             & 0.671             & 0.905             & 0.910             & \underline{0.869} & \underline{0.890}        \\
            \midrule
            \textbf{SHAMISA (ours)}                & SSL+LR
                                                   & \textbf{0.986}    & \textbf{0.987}                            & \textbf{0.981}                            & \textbf{0.987}       & \textbf{0.904}    & \textbf{0.919}    & \underline{0.922} & \underline{0.924} & \underline{0.610} & \underline{0.688} & 0.914             & \underline{0.920} & \textbf{0.886}    & \textbf{0.904}           \\
            \bottomrule
        \end{tabular}}
\end{table*}

Table~\ref{tab:main_results} shows that SHAMISA attains the best SRCC and PLCC on LIVE, CSIQ, and TID2013, while remaining competitive on KADID-10K, FLIVE, and SPAQ under the frozen-encoder linear-probe protocol.
On KADID-10K, SHAMISA ranks behind CONTRIQUE but ahead of ARNIQA. CONTRIQUE's advantage may reflect the close match between KADID-10K and the synthetic distortion families emphasized during its pre-training.
For authentic images, SHAMISA is second on FLIVE and remains competitive on SPAQ, trailing Re-IQA by a narrow margin. Together with its synthetic results, this indicates a favorable balance across the two distortion regimes.
Overall, SHAMISA achieves the strongest six-dataset average SRCC and PLCC among the SSL methods reported here, with respective averages of 0.886 and 0.904.
This balance reflects the combined use of single-factor severity trajectories and metadata-driven and structurally intrinsic relations within graph-weighted VICReg. \suppTabUncertainty{} reports splitwise variability for SHAMISA and the plain VICReg control.

These gains do not arise from a materially larger training budget. Under the same 140k-image KADIS protocol on an NVIDIA H100 80GB, the measured total times for pre-training followed by evaluation are approximately 12.5 hours for ARNIQA and 13.5 hours for SHAMISA.
Re-IQA follows a different two-encoder pipeline, training an ImageNet-based content encoder and a separate quality encoder with broad augmentation schedules.

\subsection{Cross-Dataset Transfer}
\label{subsec:cross}
\begin{table}[t]
    \centering
    \caption{\textbf{Cross-dataset SRCC on synthetic NR-IQA benchmarks.} Best and second-best results in each row are \textbf{bold} and \underline{underlined}. The comparison follows published evaluations in prior work \citep{su2020blindly,su2023distortion,madhusudana2022image,reiqa,arniqa}. We reproduce the ARNIQA column using its official implementation. The remaining baseline columns retain the published values in the cited literature. The established source-training and target-testing protocol is described below.}
    \label{tab:cross_dataset}
    \resizebox{\linewidth}{!}{
        \begin{tabular}{llcccccc}
            \toprule
            \multicolumn{1}{c}{Training} & \multicolumn{1}{c}{Testing} & HyperIQA       & Su et al.         & CONTRIQUE         & Re-IQA            & ARNIQA            & \textbf{SHAMISA}  \\
            \midrule
            LIVE                         & CSIQ                        & 0.744          & 0.777             & 0.803             & 0.795             & \underline{0.904} & \textbf{0.909}    \\
            LIVE                         & TID2013                     & 0.541          & 0.561             & 0.640             & 0.588             & \underline{0.697} & \textbf{0.700}    \\
            LIVE                         & KADID-10K                   & 0.492          & 0.506             & \underline{0.699} & 0.557             & \textbf{0.764}    & 0.694             \\
            CSIQ                         & LIVE                        & 0.926          & \underline{0.930} & 0.912             & 0.919             & 0.921             & \textbf{0.939}    \\
            CSIQ                         & TID2013                     & 0.541          & 0.550             & 0.570             & 0.575             & \underline{0.721} & \textbf{0.729}    \\
            CSIQ                         & KADID-10K                   & 0.509          & 0.515             & 0.696             & 0.521             & \textbf{0.735}    & \underline{0.711} \\
            TID2013                      & LIVE                        & 0.876          & 0.892             & \underline{0.904} & 0.900             & 0.869             & \textbf{0.908}    \\
            TID2013                      & CSIQ                        & 0.709          & 0.754             & 0.811             & 0.850             & \underline{0.866} & \textbf{0.870}    \\
            TID2013                      & KADID-10K                   & 0.581          & 0.554             & 0.640             & 0.636             & \underline{0.726} & \textbf{0.779}    \\
            KADID-10K                    & LIVE                        & \textbf{0.908} & 0.896             & 0.900             & 0.892             & 0.898             & \underline{0.901} \\
            KADID-10K                    & CSIQ                        & 0.809          & 0.828             & 0.773             & 0.855             & \underline{0.882} & \textbf{0.892}    \\
            KADID-10K                    & TID2013                     & 0.706          & 0.687             & 0.612             & \underline{0.777} & 0.760             & \textbf{0.779}    \\
            \bottomrule
        \end{tabular}
    }
\end{table}

\paragraph{Cross-dataset transfer}
Table~\ref{tab:cross_dataset} shows that SHAMISA attains the best SRCC on 9 of the 12 synthetic transfer directions, ranks second on 2 more, and remains competitive on the remaining case.
Relative to ARNIQA, gains appear in 10 of 12 directions and are largest for TID2013$~\!\rightarrow~$KADID-10K. Improvements also occur when LIVE or CSIQ transfers to TID2013. These results are consistent with varied distortion structure remaining accessible to a source-trained linear regressor. The exceptions are LIVE$~\!\rightarrow~$KADID-10K and CSIQ$~\!\rightarrow~$KADID-10K. KADID-10K$~\!\rightarrow~$LIVE is another difficult direction, where HyperIQA remains strongest and SHAMISA ranks second. LIVE is smaller and contains perceptually milder synthetic distortions, which may leave less headroom in this transfer direction.

\paragraph{Protocol}
We follow the standard SSL-IQA linear-probe setting \citep{madhusudana2022image,reiqa,arniqa}. We freeze the encoder, fit a ridge head on each source training split, select regularization on source validation data, and score the target test split with the same multi-scale five-crop features. Within each transfer direction, no target labels or target statistics enter ridge fitting, normalization, or adaptation. We report the median over 10 reference-disjoint splits, and the pre-trained configuration remains fixed throughout. This setting mirrors practical deployment, where unlabeled data support self-supervised pre-training and a limited labeled source dataset fits a lightweight head that must transfer to a related target without target-label tuning, target-specific normalization, or other target-side adaptation.
Full details are provided in \suppAppCross.
We further evaluate synthetic-to-authentic and authentic-to-authentic transfer for SHAMISA and ARNIQA in \suppAppTransfer{} and \suppTabTransfer{}. The fixed SHAMISA encoder leads ARNIQA in six of eight synthetic-to-authentic directions and both FLIVE--SPAQ directions. Every regressor setting is selected using source data only.

\subsection{gMAD Diagnostic on Waterloo}
\label{subsec:gmad}
\begin{figure}[t]
    \centering
    \includegraphics[trim=0 22 0 0,clip,width=0.24\linewidth]{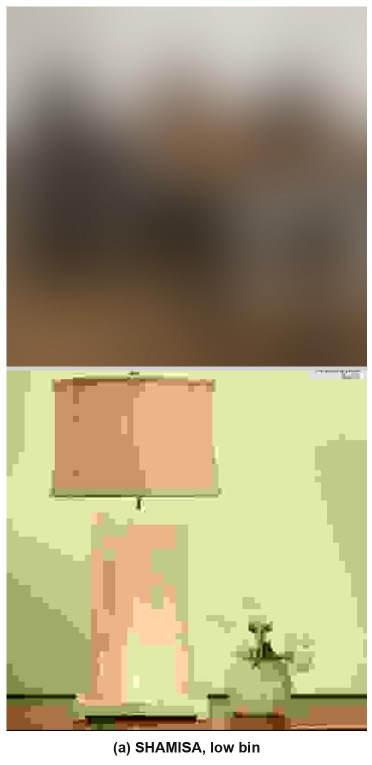}\hfill
    \includegraphics[trim=0 22 0 0,clip,width=0.24\linewidth]{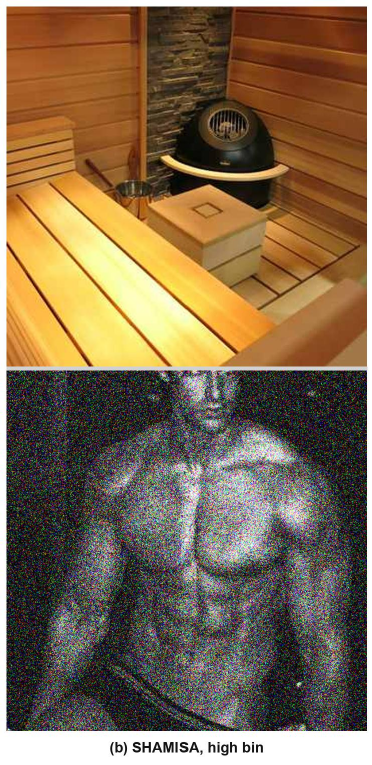}\hfill
    \includegraphics[trim=0 22 0 0,clip,width=0.24\linewidth]{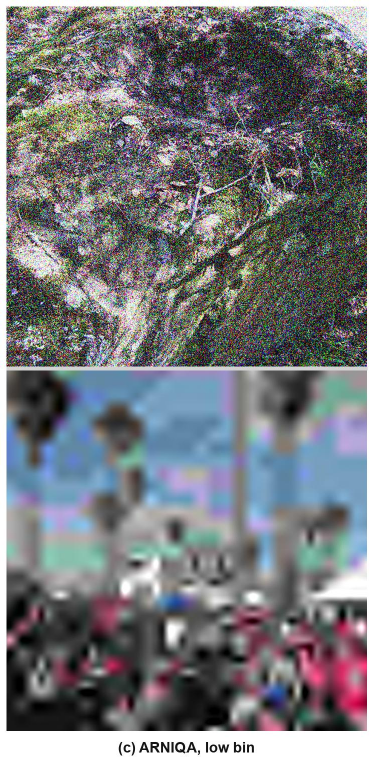}\hfill
    \includegraphics[trim=0 22 0 0,clip,width=0.24\linewidth]{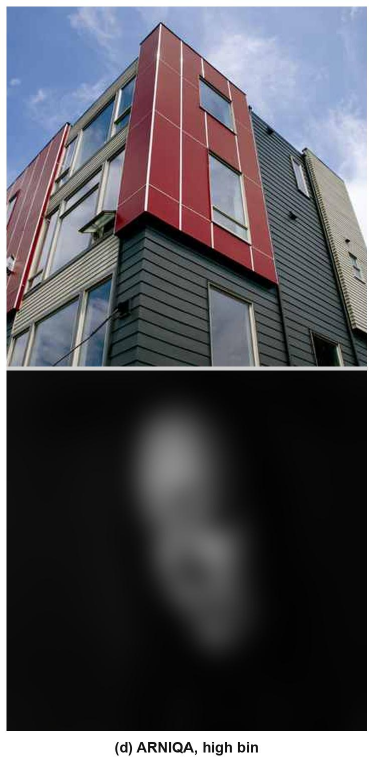}
    \par\vspace{0.35ex}
    \parbox[t]{0.24\linewidth}{\centering\small (a) SHAMISA, low bin}\hfill
    \parbox[t]{0.24\linewidth}{\centering\small (b) SHAMISA, high bin}\hfill
    \parbox[t]{0.24\linewidth}{\centering\small (c) ARNIQA, low bin}\hfill
    \parbox[t]{0.24\linewidth}{\centering\small (d) ARNIQA, high bin}
    \caption{\textbf{gMAD on the Waterloo Exploration distorted pool.} Panels (a,b) use SHAMISA as the defender and panels (c,d) use ARNIQA as the defender. In each panel, the top and bottom images form the attacker-selected pair from the indicated defender-quality bin. Low-bin cases appear in (a,c) and high-bin cases in (b,d). \suppTabGMAD\ reports the corresponding top-1 attacker gaps.}
    \label{fig:gmad_arniqa_trained}
\end{figure}

We use the group maximum differentiation (gMAD) protocol~\citep{ma2016group} as a pairwise diagnostic on the Waterloo Exploration Database~\citep{ma2016waterloo} using its distorted image pool (no MOS annotations).
Following our standard NR-IQA evaluation pipeline, we convert frozen representations into scalar quality predictions by fitting a ridge regressor on KADID-10K and then apply the resulting predictors to all Waterloo images.
Given a defender model, we first split the pool into equal-count bins based on the defender's predicted quality. Within a chosen bin, the attacker then selects the image pair with the largest attacker-side score gap while the defender still scores the pair similarly.
\suppAppgmad\ gives the corresponding protocol details.
Fig.~\ref{fig:gmad_arniqa_trained} shows representative pairs for low and high defender levels in both defender and attacker directions, and \suppTabGMAD\ reports the corresponding top-1 attacker gaps.
In this pairwise comparison with KADID-10K-trained ridge heads, the attacker gaps are smaller when SHAMISA\ is the defender than when ARNIQA is the defender. This result suggests tighter within-bin quality consistency for SHAMISA in the reported setting.

\subsection{t-SNE Representation Visualization}
\begin{figure}[t]
    \centering
    \includegraphics[width=\linewidth]{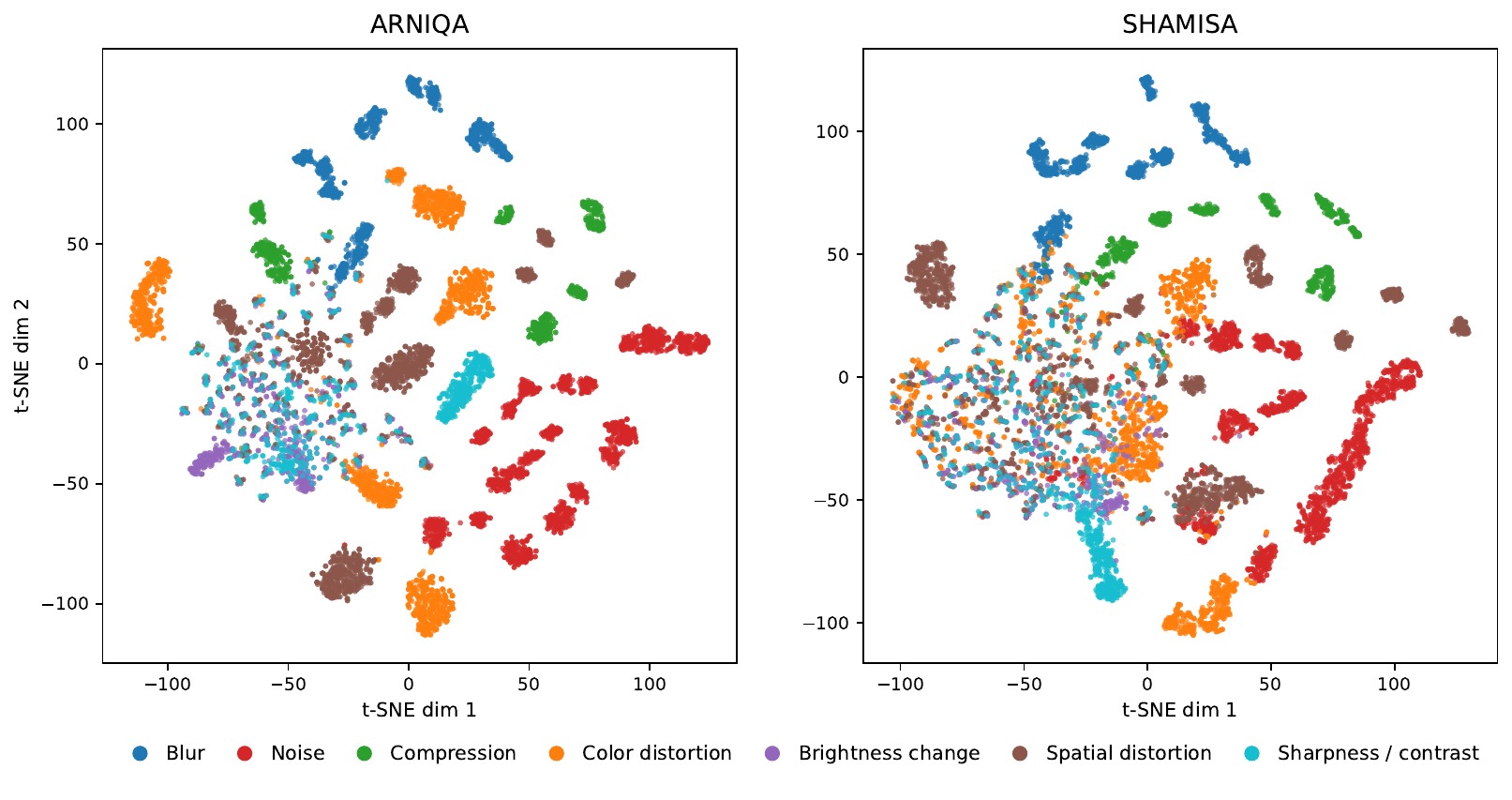}
    \caption{\textbf{t-SNE visualization of image-level encoder representations on KADID-10K.}
        We extract encoder representations $\mathbf{H}$ (pre-projector) and average across crops to obtain one representation per image, yielding 10,125 image-level points.
        Points are colored by the coarse distortion family using a shared fixed palette.
        SHAMISA shows clearer coarse-family organization for several prominent degradation families, while both models retain overlap among visually related brightness and color manipulations.
        Each panel is a fixed-seed 2D projection used as a qualitative diagnostic.}
    \label{fig:tsne_kadid_groups}
\end{figure}
We visualize the structure of the learned representation space by applying 2D t-SNE~\citep{van2008visualizing} to image-level encoder representations $\mathbf{H}$ on KADID-10K~\citep{kadid10k}.
For each image, we extract $\mathbf{H}$ (pre-projector) and average across crops, yielding $10{,}125$ points.
Fig.~\ref{fig:tsne_kadid_groups} colors points by coarse distortion family using a shared fixed palette.
Both models separate several major degradation families. SHAMISA forms more coherent regions for prominent categories in the displayed projection, while ARNIQA shows greater inter-family mixing near the center.
We provide additional severity-colored visualizations and subtype-level views (by distortion family) in \suppApptsne.
Across these plots, separation is clearest for families with distinctive artifact statistics such as blur, noise, and compression. Brightness-related manipulations remain more entangled for both models. These qualitative observations are consistent with SHAMISA encouraging coherent organization across major distortion families while retaining variation within them.

\subsection{UMAP Manifold Visualization}
\begin{figure}[t]
    \centering
    \subfloat[ARNIQA\label{fig:umap_manifold_arniqa_trained}]{%
        \includegraphics[width=0.49\linewidth]{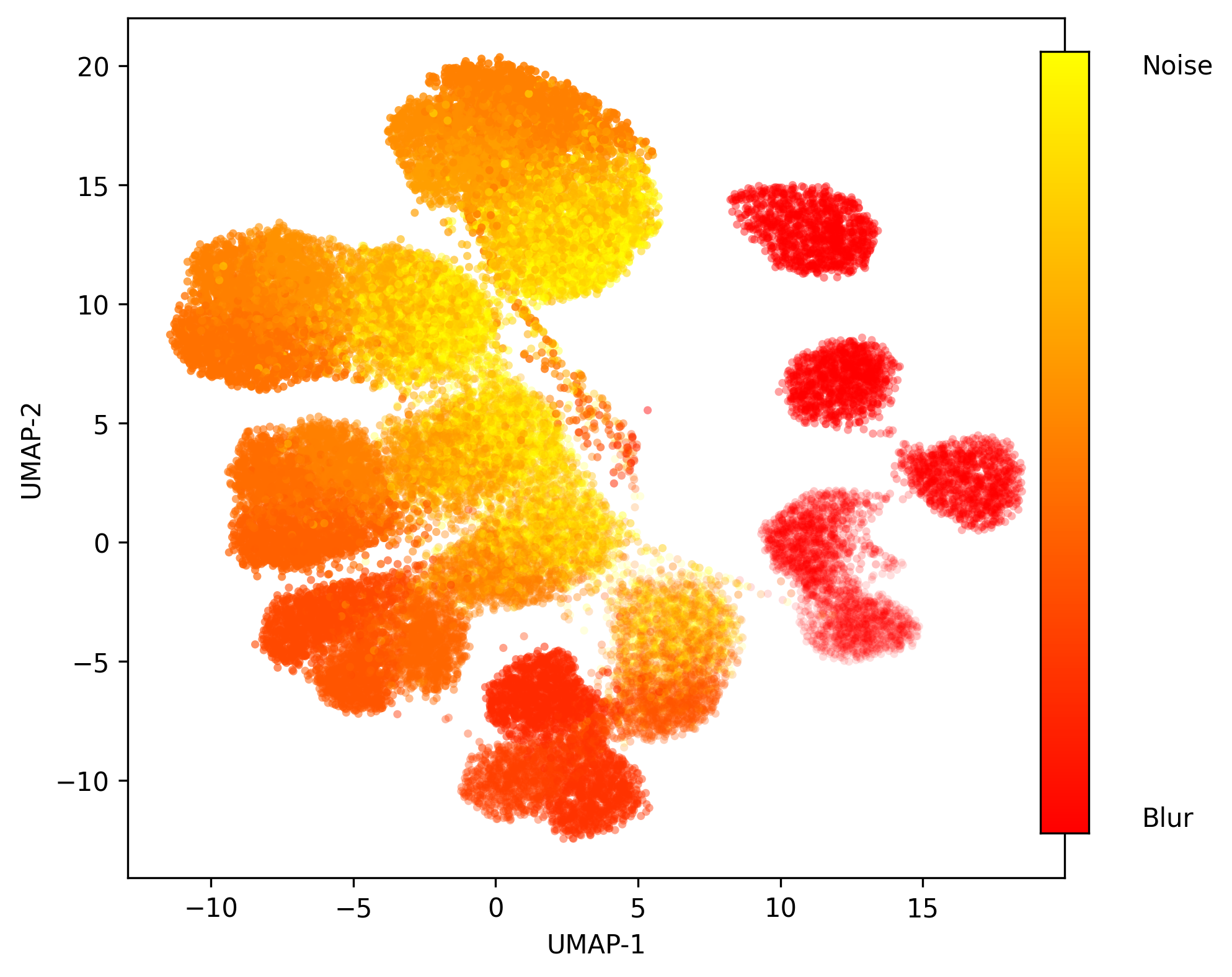}}
    \hfill
    \subfloat[SHAMISA\label{fig:umap_manifold_shamisa}]{%
        \includegraphics[width=0.49\linewidth]{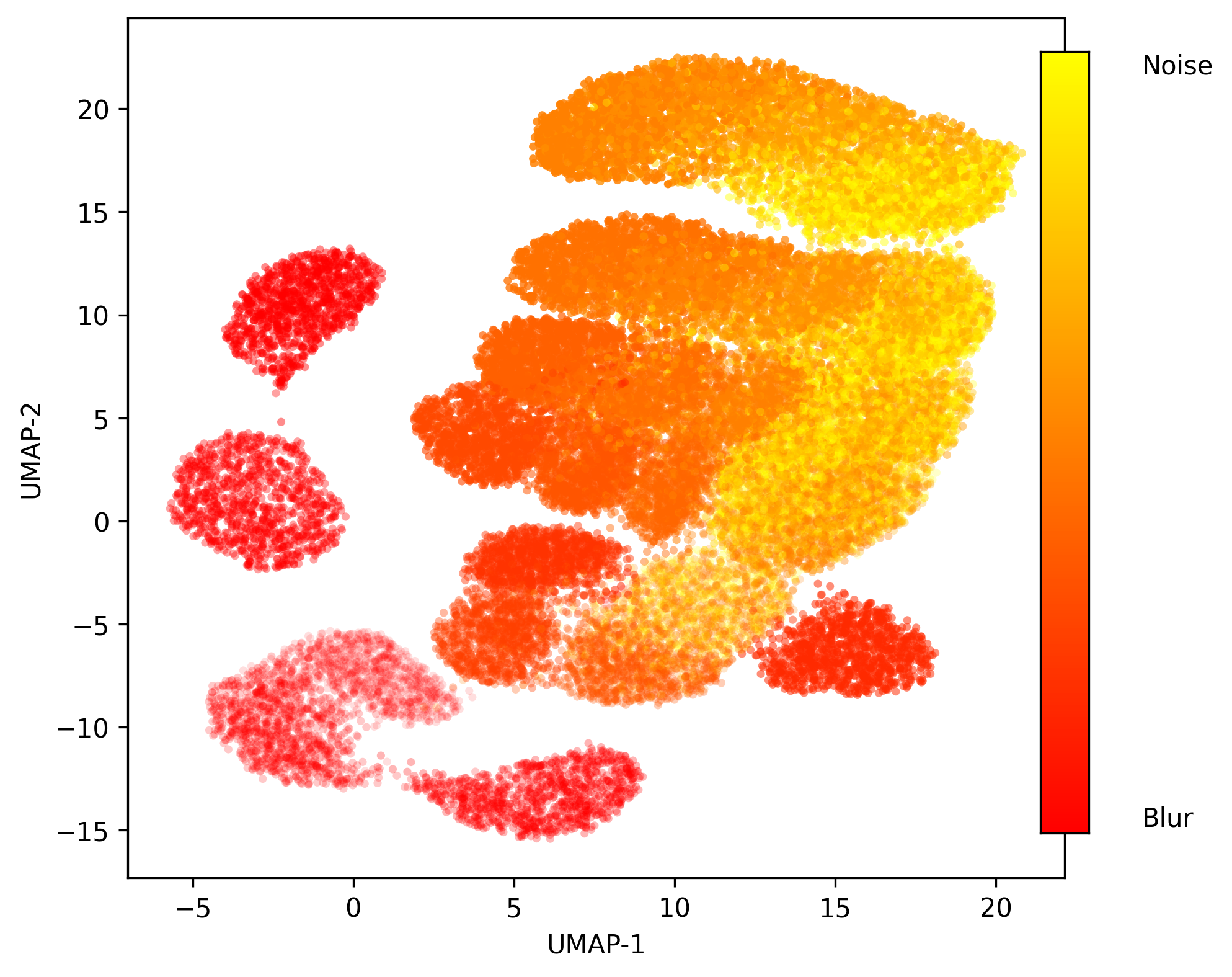}}
    \caption{
        \textbf{Manifold visualization with UMAP \citep{mcinnes2018umap}} using encoder representations $\mathbf{H}$ extracted from 1,000 pristine KADIS images degraded with Gaussian blur and white noise.
        For each pristine image we generate 5 blur-only samples, 5 noise-only samples, and all $5 \times 5$ blur$\rightarrow$noise compositions (35 variants per image, 35k total points).
        Point color is the weighted average of blur in red and noise in yellow according to their relative intensities. Marker opacity increases with total severity.
    }
    \label{fig:umap_manifold}
\end{figure}

\label{sec:umap_manifold}

We visualize the distortion manifold induced by two degradations and their compositions using UMAP \citep{mcinnes2018umap}.
We sample 1,000 pristine images from KADIS and generate 35 degraded variants per image: 5 blur-only samples, 5 noise-only samples, and all $5 \times 5$ blur$\rightarrow$noise compositions, where blur is applied before noise.
We then extract pooled encoder features $\mathbf{H}$ (pre-projector) with frozen weights and embed them into 2D using identical UMAP hyperparameters across models (Fig.~\ref{fig:umap_manifold}). Protocol details are given in \suppAppumap.
We compare the internal organization within each independently fitted panel.

Across both models, compositions tend to lie closer to the noise region, which is consistent with noise being applied after blur and remaining visually prominent in the final image. The ARNIQA embedding contains a diffuse interior region where different blur-to-noise mixtures overlap. SHAMISA shows clearer bands and a smoother progression from blur-dominated to noise-dominated compositions, with less central mixing. This internal organization suggests that SHAMISA captures the two distortion factors more coherently in this diagnostic.

\subsection{Ablations}
\label{subsec:ablations}
\begin{table}[t]
    \centering
    \scriptsize
    \setlength{\tabcolsep}{4.2pt}
    \caption{\scriptsize
        \textbf{Main ablation results.} Each dataset contributes its median across the available splits (10 where applicable; FLIVE uses its official split).
        Subscripts $s$, $a$, and $g$ denote averages of these per-dataset medians over the synthetic datasets, authentic datasets, and all six datasets, respectively.
        The complete catalog is provided in \suppTabAbl.}
    \label{tab:ablations_main}
    \resizebox{\columnwidth}{!}{%
        \begin{tabular}{lcccccc}
            \toprule
            \textbf{Variant}                                   &
            \textbf{SRCC$_g$}                                  & \textbf{PLCC$_g$} &
            \textbf{SRCC$_a$}                                  & \textbf{PLCC$_a$} &
            \textbf{SRCC$_s$}                                  & \textbf{PLCC$_s$}                                              \\
            \midrule
            \textbf{A0: SHAMISA}                               & 0.8862            & 0.9042 & 0.7620 & 0.8040 & 0.9483 & 0.9543 \\
            \midrule
            \multicolumn{7}{l}{\textit{(A) Aggregation \& regularization}}                                                      \\
            A1: Binary $\mG$                                   & 0.8767            & 0.8924 & 0.7475 & 0.7821 & 0.9374 & 0.9457 \\
            A3: No $\Phi$ (fixed linear mixture)               & 0.8759            & 0.8951 & 0.7479 & 0.7896 & 0.9361 & 0.9457 \\
            \midrule
            \multicolumn{7}{l}{\textit{(B) Graph sources}}                                                                      \\
            B1: No $\mG^{o}$ (OT graph)                        & 0.8600            & 0.8787 & 0.7661 & 0.8069 & 0.8995 & 0.9092 \\
            B7: Metadata only $\{\mG^{rd},\mG^{dd},\mG^{rr}\}$ & 0.8467            & 0.8648 & 0.7634 & 0.8010 & 0.8801 & 0.8905 \\
            B6: $\mG^{k}$ only                                 & 0.7891            & 0.8018 & 0.7577 & 0.7825 & 0.7922 & 0.8017 \\
            B8: Structure only $\{\mG^{k},\mG^{o}\}$           & 0.8753            & 0.8961 & 0.7419 & 0.7891 & 0.9387 & 0.9475 \\
            \midrule
            \multicolumn{7}{l}{\textit{(D) Distortion engine}}                                                                  \\
            D1: $M_d{=}1$                                      & 0.8775            & 0.9077 & 0.7259 & 0.7684 & 0.9519 & 0.9783 \\
            D2: Discrete severities ($L{=}5$)                  & 0.8767            & 0.8987 & 0.7456 & 0.7890 & 0.9386 & 0.9518 \\
            D4: Fixed composition order $\pi$                  & 0.8587            & 0.8712 & 0.7434 & 0.7668 & 0.9113 & 0.9215 \\
            \midrule
            \multicolumn{7}{l}{\textit{(E) Objective diagnostic}}                                                               \\
            E1: No $\mathcal{L}_{\mathrm{ot}}$                 & 0.7887            & 0.8009 & 0.7597 & 0.7836 & 0.7904 & 0.7996 \\
            \bottomrule
        \end{tabular}
    }
\end{table}

\paragraph{Ablation design}
We organize the variants by the design question they test: (A) aggregation and regularization, (B) retained relation sources, (C) OT-graph sparsification, (D) the compositional training distribution, (E) objective form, and (F) severity mapping. Each variant changes a minimal set of factors relative to A0. Table~\ref{tab:ablations_main} highlights the most diagnostic variants across LIVE, CSIQ, TID2013, KADID-10K, FLIVE, and SPAQ. \suppTabAbl\ provides the complete A--F catalog.

\begin{sloppypar}
    \hbadness=10000
    \paragraph{Results and analysis}
    A0 achieves SRCC$_g{=}0.8862$ (PLCC$_g{=}0.9042$), with SRCC$_s{=}0.9483$ and SRCC$_a{=}0.7620$.
    Across the complete catalog in \suppTabAbl, three trends support the design of SHAMISA. First, OT-guided structure makes an important contribution. Removing the OT alignment term in E1 yields the largest decrease to SRCC$_g{=}0.7887$, while removing the OT-derived graph in B1 produces the strongest decrease among the single-source removals to 0.8600. These results indicate that long-range OT relations complement local invariance.
\end{sloppypar}

Second, metadata-driven and structurally intrinsic relations play complementary roles. The local-structure variant B6 yields SRCC$_g{=}0.7891$, while metadata-only B7 reaches 0.8467. This pattern is consistent with structural graphs preserving learned neighborhoods while metadata graphs provide distortion information. Structure-only B8 remains competitive on synthetic data (SRCC$_s{=}0.9387$) but decreases to SRCC$_a{=}0.7419$ on authentic data. The modest changes from individual metadata-source removals B3--B5 further indicate that these relations are most useful collectively.

Third, the distortion engine benefits from composition diversity and continuous severity modeling. Restricting each composition to one atomic distortion in D1 preserves strong synthetic performance but lowers authentic SRCC from 0.7620 to 0.7259. This result indicates that exposure to richer compositions is useful for authentic datasets, where multiple degradations can coexist. Fixing the composition order in D4 lowers SRCC$_g$ to 0.8587, showing that randomized order contributes useful diversity. Discrete severity levels in D2 reduce SRCC$_g$ to 0.8767, while uniform severity sampling in D3 gives 0.8743. These results support continuous severity sampling for fine-grained trajectories with retained exposure to severe distortions.

Finally, the aggregation variants show the value of graded and adaptive relations. Binarizing $\mG$ in A1 reduces SRCC$_g$ to 0.8767, while replacing $\Phi$ with a fixed linear mixture in A3 gives 0.8759. Removing the graph regularizer in A2 produces a smaller decrease to 0.8813, which suggests a supporting rather than dominant role.

\subsection{Pre-training Dynamics}
\label{subsec:dynamics}
\label{sec:training_dynamics}
\begin{figure}[t]
    \centering
    \includegraphics[width=\linewidth]{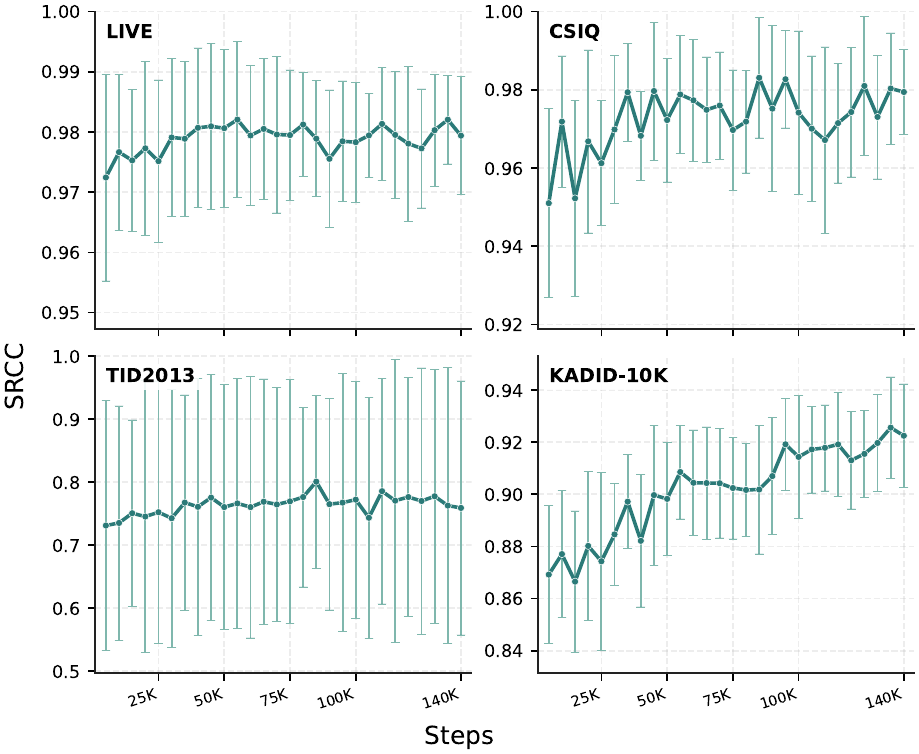}
    \vspace{\figup}
    \caption{
        \textbf{SRCC vs. training steps during SSL pre-training} (mean with $\pm$ std error bars over 10 splits) across representative NR-IQA benchmarks.
        Curves typically rise early and then saturate, while the best checkpoint can vary by dataset because the self-supervised objective differs from downstream SRCC.
    }
    \label{fig:srcc_vs_steps}
    \vspace{\figdown}
\end{figure}

We track downstream NR-IQA performance during SSL pre-training by periodically freezing the encoder and fitting a lightweight regressor on each target dataset.
Frozen checkpoints are evaluated every 5000 training steps.
Fig.~\ref{fig:srcc_vs_steps} shows that SRCC improves rapidly in the early phase and then largely saturates across the representative benchmarks, indicating that the learned representation becomes useful for quality prediction relatively early. Importantly, the best checkpoint can be dataset-dependent: some datasets continue to improve or remain stable, while others show a mild peak and decline. This is expected because the self-supervised objective is not optimized directly for SRCC on any one benchmark.
As a deployment note, when a small labeled validation set is available in the target domain, it can be used to select a checkpoint for that domain.
Unless stated otherwise, all results reported in Tables~\ref{tab:main_results} and~\ref{tab:cross_dataset} use the fixed A0 checkpoint. This section is a diagnostic and is not used for model selection.
Additional diagnostic figures are provided in \suppAppDynamics.

\subsection{Hyperparameter Sensitivity}
\label{subsec:hp}
\label{sec:hp_sensitivity}

\begin{figure}[t]
    \centering
    \includegraphics[width=\linewidth]{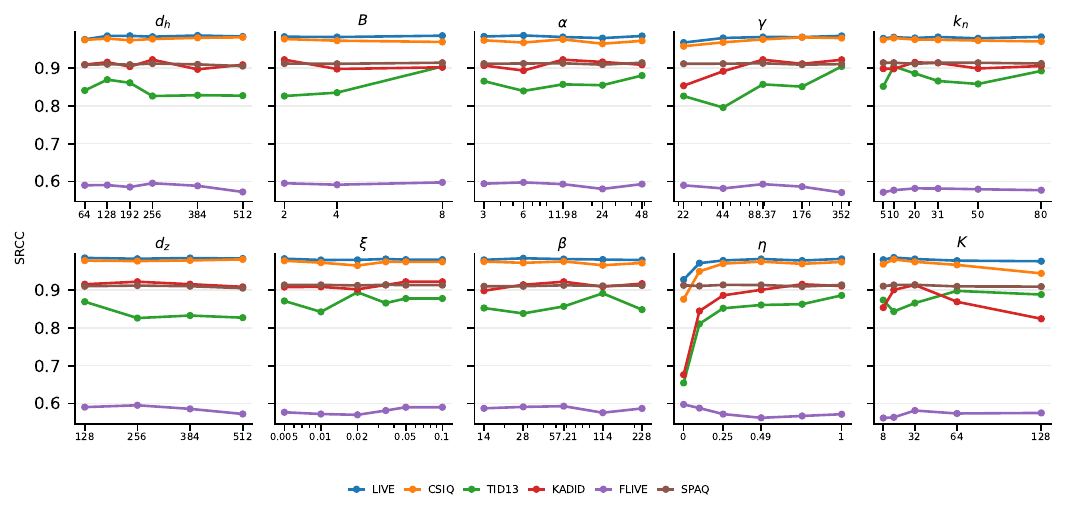}
    \caption{
        \textbf{One-at-a-time sweeps around the selected SHAMISA configuration.} The panels vary $(d_h,d_z)$, $(B,\xi)$, $(\alpha,\beta)$, $(\gamma,\eta)$, and $(k_n,K)$ while reporting SRCC across six NR-IQA benchmarks.
    }
    \label{fig:hp_sensitivity_merged}
\end{figure}

We analyze hyperparameter sensitivity by varying one knob at a time around the selected configuration while keeping the rest fixed.
Fig.~\ref{fig:hp_sensitivity_merged} shows broad performance plateaus for most capacity and graph-construction choices. Projector width and $k_n$ yield relatively stable SRCC across their tested ranges, while increasing $K$ generally saturates after moderate values. The OT alignment coefficient $\eta$ shows the strongest variation, especially on TID2013 and KADID-10K. This pattern suggests that weak alignment may underuse relational guidance, while excessive alignment can constrain the representation. Among the VICReg terms, $\gamma$ is the most sensitive, with its largest ranges on TID2013 and KADID-10K. On FLIVE, its effect approaches that of $\eta$, while $\alpha$ and $\beta$ are more stable. Overall, performance is comparatively stable over most tested ranges, with $\eta$ and $\gamma$ providing finer control. \suppAppHPSens\ reports the corresponding SRCC summary.

\vspace{0.5ex}
\paragraph{FR extension}
We also propose a full-reference variant, SHAMISA-FR, that reuses the frozen no-reference encoder and uses the absolute feature difference \(|h_{\mathrm{ref}}{-}h_{\mathrm{dist}}|\) between pristine and distorted images as the regressor input. This variant requires no additional backbone pre-training and attains competitive FR-IQA accuracy. Details are provided in \suppAppFR, with the main comparison in \suppTabFR.

\vspace{0.5ex}
\paragraph{Reproducibility}
Additional implementation details, optimization settings, evaluation protocols, and supplementary diagnostics are provided in \suppAppArchitectures, \suppAppCross, \suppAppDynamics, and \suppAppHPSens. The project repository is available at \url{https://github.com/Mahdi-Naseri/SHAMISA/}.

\section{Conclusion}
\label{sec:conclusion}
We presented \textbf{SHAMISA}, a non-contrastive self-supervised framework for NR-IQA that learns unified distortion-aware and content-aware representations. SHAMISA couples a compositional distortion engine with single-factor variation to generate controllable severity trajectories and supervises learning through dual-source relation graphs that fuse Metadata-Driven Graphs with Structurally Intrinsic Graphs. These relations drive graph-weighted VICReg-style invariance with clustering-guided supervision. In each iteration, the graphs are constructed from the current representations, treated with stop-gradient, and then used to update the model. The resulting frozen encoder supports accurate quality prediction with a linear regressor. Across synthetic and authentic benchmarks, SHAMISA achieves the strongest overall average among the compared SSL methods and improves transfer in most reported synthetic cross-dataset directions. The synthetic-to-authentic and authentic-to-authentic experiments further demonstrate fixed-configuration transfer to held-out authentic target test images without per-target tuning. The gMAD analysis provides complementary qualitative evidence in its stated setting. A full-reference variant, SHAMISA-FR, reuses the frozen encoder without additional backbone pre-training and attains competitive FR-IQA accuracy. Future work includes exploiting the learned topology for blind image restoration and extending the framework to video quality assessment.

\ifshamisacombinedbuild
\else
    \section*{Supplementary Material}
    The appendices referenced throughout this manuscript are provided in the accompanying supplementary material.

    \bibliographystyle{IEEEtran}
    \bibliography{refs}
\fi

%% file: main.bbl

%% file: 1_supplementary.tex
\appendices

\section{Architectures and Optimization}
\label{app:exp-details}
Table~\ref{tab:besthp} summarizes the fixed SHAMISA pre-training configuration used across all experiments. Unless stated otherwise, downstream evaluation selects the ridge regularization \(\alpha_{\mathrm{ridge}}\) per dataset on the validation split as described in Sec.~\ref{subsec:datasets-protocol}.

\begin{table*}[t]
    \centering
    \footnotesize
    \caption{\textbf{Essential final hyperparameters used for SHAMISA pre-training.} A single configuration is shared across all datasets (no dataset-specific tuning during SSL).}
    \label{tab:besthp}
    \setlength{\tabcolsep}{3pt}
    \begin{tabular}{@{}lp{0.84\textwidth}@{}}
        \toprule
        Group             & Hyperparameters                                                                                                                                          \\
        \midrule
        Architecture      &
        ResNet-50 encoder with $d_h=2048$ and projector output $d_z=256$.                                                                                                            \\
        Optimization      &
        SGD with $\mathrm{lr}=1.5\times10^{-3}$, momentum $0.9$, weight decay $10^{-4}$, and cosine annealing warm restarts.                                                         \\
        Loss              &
        VICReg-style weights $(\alpha,\beta,\gamma)=(11.98, 57.21, 88.37)$, OT alignment weight $\eta=0.4906$, and graph regularizer weight $\xi=0.0342$.                           \\
        Graphs            &
        kNN neighborhood size $k_n=31$, metadata sparsifier $K_d=4096$, reference--reference weight $w_{rr}=0.5766$, OT prototypes $K=32$, temperature $\tau_c=0.1$, six Sinkhorn iterations with entropy parameter $0.05$, and OT sparsifier $K_g=8N$. \\
        Distortion engine &
        Each update uses $B=2$ tiny-batches, $R=3$ references, $C=4$ composition groups, $L=5$ severity levels, crop size $224$, and at most $M_d=7$ distortions per sample.       \\
        \bottomrule
    \end{tabular}
\end{table*}

\noindent\emph{Batch construction details.}
For each update, one random crop per pristine image forms
\[
    \mR^s \in \R^{N_{\mathrm{ref}}\times 3\times H_{in}\times W_{in}}.
\]
For $d=1,\dots,N_{\mathrm{comp}}$, $\mX^{(d)}=\chi_d(\mR^s)$ denotes the degraded crop block associated with composition index $d$ and has the same tensor shape as $\mR^s$. Each tiny-batch uses its own sampled compositions. Concatenating $\mR^s$ and all $\mX^{(d)}$ gives the batch $\mX$ in Eq.~\ref{eq:batch_concat}. The encoder and projector then produce $\mH\in\R^{N\times d_h}$ and $\mZ\in\R^{N\times d_z}$.

Following the validation-based selection protocol used in prior work~\citep{madhusudana2022image,reiqa,arniqa}, the final configuration was selected through Bayesian optimization with Weights \& Biases~\citep{wandb}. The selection criterion was the average validation SRCC across LIVE, CSIQ, TID2013, KADID-10K, FLIVE, and SPAQ. Repeated validation splits were used where available, while FLIVE followed its official split. The sweep covered the loss coefficients, representation dimensions, graph construction settings, sparsification, and regularization. The selected A0 configuration gave the highest aggregate validation performance and was then fixed for all reported evaluations.

Before aggregation, we remove diagonal self-edges from every source graph. Consequently, Eq.~\ref{eq:g_inv} contains no self-invariance terms.

\section{Compositional Distortion Engine}
We use the compositional engine of Sec.~\ref{sec:distortion_engine} with $M_d=7$ and $B=2$ tiny-batches per update. Each tiny-batch contains $R=3$ pristine references and $C=4$ distortion composition groups. Every group forms a single-factor trajectory with $L=5$ levels and generates images $x_{i,j,k,l}$ as defined in Eq.~\ref{eq:engine_gen}. Normalized severities are sampled independently by drawing $\epsilon\sim\mathcal{N}(0,0.5^2)$ and setting $s=\min(1,|\epsilon|)$. The distribution satisfies $\Pr(s<0.5)=68.27\%$ and $\Pr(s=1)=4.55\%$, so it emphasizes mild severity while retaining maximum-severity samples. The uniform-sampling ablation D3 provides an empirical comparison with this choice. Within each group, one distortion coordinate varies across $L$ independent draws $\{s^{(l)}\}_{l=1}^{L}$ that remain in sampled order, while all other coordinates retain their sampled baseline values. Distortion functions are applied sequentially in a uniformly random order according to Eq.~\ref{eq:composition}.

\section{Evaluation Details}
Following prior SSL-IQA evaluation practice \citep{madhusudana2022image,reiqa,arniqa}, we use five deterministic crop positions at full and half image scales \citep{madhusudana2022image,zhao2023quality}. At each position, we concatenate the two scale features, apply the ridge head, and then average the five scalar predictions. Ridge regularization is selected on the validation split as described in Sec.~\ref{subsec:datasets-protocol}. We report the median over 10 random splits where applicable. Synthetic datasets use reference-disjoint splits following \citep{arniqa,madhusudana2022image,reiqa}. FLIVE uses its official split \citep{ying2020patches}, while SPAQ uses random 70, 10, and 20 percent training, validation, and test partitions. PLCC follows the four-parameter logistic mapping specified in Sec.~\ref{subsec:datasets-protocol} for within-dataset evaluation. The transfer analysis reports SRCC only because fitting the usual PLCC mapping on target labels would leak target information.

\section{Datasets}
\label{app:data}
We evaluate on four synthetic IQA benchmarks and two authentic in-the-wild IQA benchmarks. Table~\ref{tab:dataset_stats_appendix} summarizes the dataset sizes, label types, and split notes used throughout the paper.

Among the authentic datasets, FLIVE provides large-scale in-the-wild images with both image-level and patch-level quality annotations \citep{ying2020patches}, whereas SPAQ focuses on smartphone photography captured under diverse real-world conditions \citep{fang2020perceptual}. For SPAQ, we resize the shorter side to 512 as in \citep{fang2020perceptual}.

\begin{table}[t]
    \centering
    \scriptsize
    \setlength{\tabcolsep}{3pt}
    \renewcommand{\arraystretch}{1.05}
    \caption{\textbf{Dataset statistics for the NR-IQA benchmarks.} Synthetic datasets are reported by reference count and distortion inventory, whereas authentic datasets are summarized by image count and label type.}
    \label{tab:dataset_stats_appendix}
    \resizebox{\columnwidth}{!}{
        \begin{tabular}{lcccccc}
            \toprule
            Dataset                              & Type        & Pristine refs & Distortion types & Images   & Label type         & Split note         \\
            \midrule
            LIVE \citep{sheikh2006statistical}   & Synthetic   & 29            & 5                & 779      & DMOS               & Reference-disjoint \\
            CSIQ \citep{larson2010most}          & Synthetic   & 30            & 6                & 866      & DMOS               & Reference-disjoint \\
            TID2013 \citep{ponomarenko2013color} & Synthetic   & 25            & 24               & 3000     & MOS                & Reference-disjoint \\
            KADID-10K \citep{kadid10k}           & Synthetic   & 81            & 25               & 10{,}125 & MOS                & Reference-disjoint \\
            FLIVE \citep{ying2020patches}        & In-the-wild & --            & --               & 39{,}808 & MOS + patch scores & Official split     \\
            SPAQ \citep{fang2020perceptual}      & In-the-wild & --            & --               & 11{,}125 & MOS                & Random split       \\
            \bottomrule
        \end{tabular}
    }
\end{table}

\section{SHAMISA-FR: Full-Reference Variant}
\label{app:fr}
\begin{table*}[t]
    \centering
    \scriptsize
    \setlength{\tabcolsep}{4.0pt}
    \renewcommand{\arraystretch}{1.05}
    \caption{\textbf{Full-reference IQA comparison on synthetic distortion datasets.}
        Reported rows are taken from prior work. SHAMISA-FR uses the same fixed-encoder evaluation protocol as our NR setting, and each value is the median over 10 random splits. Best and second-best values in each column are shown in bold and underlined.}
    \label{tab:fr_iqa_synthetic}
    \resizebox{\textwidth}{!}{
        \begin{tabular}{l l cc cc cc cc}
            \toprule
            Method                                    & Type
                                                      & \multicolumn{2}{c}{\live}
                                                      & \multicolumn{2}{c}{\csiq}
                                                      & \multicolumn{2}{c}{\tid}
                                                      & \multicolumn{2}{c}{\kadid}                                                                                                                                                     \\
                                                      &                            & \srcc             & \plcc             & \srcc             & \plcc             & \srcc             & \plcc             & \srcc             & \plcc \\
            \midrule
            PSNR                                      & Traditional
                                                      & 0.881                      & 0.868             & 0.820             & 0.824             & 0.643             & 0.675             & 0.677             & 0.680                     \\
            SSIM \citep{wang2004image}                & Traditional
                                                      & 0.921                      & 0.911             & 0.854             & 0.835             & 0.642             & 0.698             & 0.641             & 0.633                     \\
            FSIM \citep{zhang2011fsim}                & Traditional
                                                      & 0.964                      & 0.954             & 0.934             & 0.919             & 0.852             & 0.875             & 0.854             & 0.850                     \\
            VSI \citep{zhang2014vsi}                  & Traditional
                                                      & 0.951                      & 0.940             & 0.944             & 0.929             & 0.902             & 0.903             & 0.880             & 0.878                     \\
            \midrule
            PieAPP \citep{prashnani2018pieapp}        & Deep learning
                                                      & 0.915                      & 0.905             & 0.900             & 0.881             & 0.877             & 0.850             & 0.869             & 0.869                     \\
            LPIPS \citep{zhang2018unreasonable}       & Deep learning
                                                      & 0.932                      & 0.936             & 0.884             & 0.906             & 0.673             & 0.756             & 0.721             & 0.713                     \\
            DISTS \citep{ding2020image}               & Deep learning
                                                      & 0.953                      & 0.954             & 0.942             & 0.942             & 0.853             & 0.873             & --                & --                        \\
            DRF-IQA \citep{kim2020dynamic}            & Deep learning
                                                      & \textbf{0.983}             & \textbf{0.983}    & 0.964             & 0.960             & \textbf{0.944}    & \textbf{0.942}    & --                & --                        \\
            \midrule
            CONTRIQUE-FR \citep{madhusudana2022image} & SSL + LR
                                                      & 0.966                      & 0.966             & 0.956             & 0.964             & 0.909             & 0.915             & \textbf{0.946}    & \textbf{0.947}            \\
            Re-IQA-FR \citep{reiqa}                   & SSL + LR
                                                      & \underline{0.969}          & \underline{0.974} & 0.961             & 0.962             & \underline{0.920} & \underline{0.921} & \underline{0.933} & \underline{0.936}         \\
            ARNIQA-FR \citep{arniqa}                  & SSL + LR
                                                      & \underline{0.969}          & 0.972             & \textbf{0.971}    & \underline{0.975} & 0.898             & 0.901             & 0.920             & 0.919                     \\
            \textbf{SHAMISA-FR}                       & SSL + LR
                                                      & 0.962                      & 0.970             & \underline{0.968} & \textbf{0.982}    & 0.909             & 0.906             & 0.899             & 0.892                     \\
            \bottomrule
        \end{tabular}
    }
\end{table*}

\paragraph{Setup}
SHAMISA-FR reuses the pre-trained $f_{\theta}$ from the no-reference setting and keeps it frozen. The FR variant adds only a lightweight regressor, so it requires no additional backbone pre-training and only minimal extra computation.
Given $(x_{\mathrm{ref}}, x_{\mathrm{dist}})$, we compute
\[
    h_{\mathrm{ref}} = f_{\theta}(x_{\mathrm{ref}}),
    \qquad
    h_{\mathrm{dist}} = f_{\theta}(x_{\mathrm{dist}}),
\]
then form $u=\lvert h_{\mathrm{ref}}-h_{\mathrm{dist}}\rvert$.
A linear regressor is trained on $u$ to predict the quality label. LIVE and CSIQ use DMOS, while TID2013 and KADID-10K use MOS. The encoder $f_{\theta}$ remains fixed.

\paragraph{Protocol}
We evaluate on four synthetic FR-IQA datasets consisting of LIVE, CSIQ, TID2013, and KADID-10K. The evaluation uses the same split logic as the NR setting and reports median SRCC and PLCC over 10 random splits.
For context, Table~\ref{tab:fr_iqa_synthetic} also lists representative FR-IQA baselines as reported in prior work \citep{madhusudana2022image,reiqa,arniqa}.

\paragraph{Results}
Table~\ref{tab:fr_iqa_synthetic} shows that SHAMISA-FR is most competitive on CSIQ among the SSL-based FR rows, while remaining competitive on LIVE and trailing more clearly on TID2013 and KADID-10K. Unlike supervised FR baselines that optimize complete networks on target datasets, SHAMISA-FR tests a frozen no-reference representation with only a lightweight head. The results therefore reflect the quality information retained by the pre-trained representation.

\section{Cross-Dataset Evaluation Protocol}
\label{app:cross}

\paragraph{Practical motivation}
The protocol reflects a common deployment setting in which abundant unlabeled images support self-supervised pre-training, while a limited labeled source dataset is available for fitting a lightweight head. The resulting model is then applied to a related target distribution without target labels. Freezing the encoder, avoiding target-specific normalization, and selecting all regressor settings on the source data prevent target information from entering ridge training or adaptation. The A0 pre-training configuration, selected beforehand under the validation protocol in Appendix~A, remains fixed throughout these evaluations.

\paragraph{Setting}
We assess zero-shot transfer across synthetic NR-IQA datasets \{LIVE, CSIQ, TID2013, KADID-10K\}. The encoder $f_\theta$ is frozen for all evaluations and pre-trained once under the setup of Sec.~\ref{subsec:setup}. Only a ridge regressor is trained per source dataset.

\paragraph{Transfer procedure}
For each source dataset $D_s$, we create 10 reference-disjoint splits with 70 percent for training, 10 percent for validation, and 20 percent for testing. We extract frozen features at two scales and five crops per scale, concatenate the features, and train a ridge regressor on the training partition of $D_s$. The regularization coefficient is selected once on the validation partition of $D_s$ via a logarithmic grid from $10^{-3}$ to $10^{3}$. The selected regressor is then evaluated without modification on the test partition of every target $D_t \neq D_s$.

\paragraph{Preprocessing and normalization}
All synthetic datasets in this protocol use the same resizing, cropping, and color normalization pipelines. No statistics from $D_t$ are computed or applied. In particular, we do not apply target-specific normalization or retune hyperparameters.

\paragraph{Metrics and aggregation}
We report SRCC on raw predictions. For each $D_s \to D_t$, the headline result is the median over the 10 source splits.

\paragraph{Contrast to some prior practice}
Some SSL-IQA works emphasize within-dataset or mixed-dataset training. Our protocol instead evaluates transfer from $D_s$ to $D_t$ with a fixed encoder and a head selected only on $D_s$. No adaptation uses target labels or target statistics.

\section{Transfer to Authentic IQA Datasets}
\label{app:authentic_transfer}

Table~\ref{tab:authentic_transfer} evaluates transfer from synthetic and authentic source datasets to authentic targets. For every direction, the encoder is frozen, ridge regularization is selected using source validation data, and only target test images are scored. When source and target datasets use opposite MOS and DMOS conventions, prediction orientation is reversed according to the published dataset definitions without using target results. We report SRCC only. The usual nonlinear PLCC mapping cannot be fitted without target labels and would therefore introduce target-data leakage.

Each entry is the median across 10 source and target split pairings. For a synthetic source, split $i$ is paired with SPAQ target split $i$, while FLIVE always contributes its official target test split. For FLIVE$\rightarrow$SPAQ, its single official source split is paired with the 10 SPAQ target splits. For SPAQ$\rightarrow$FLIVE, the 10 SPAQ source splits are paired with the official FLIVE target test split. SHAMISA and ARNIQA~\citep{arniqa} use their designated pre-trained checkpoints under the same evaluation procedure. Every SHAMISA direction uses the fixed selected encoder and unchanged OT hyperparameters, with no retuning for an authentic target. SHAMISA leads in six of eight synthetic-to-authentic directions and in both authentic-to-authentic directions. These results demonstrate fixed-configuration transfer to held-out authentic target test images without per-target tuning.

\begin{table}[!h]
    \centering
    \scriptsize
    \setlength{\tabcolsep}{4.0pt}
    \renewcommand{\arraystretch}{1.05}
    \caption{\textbf{Transfer to authentic IQA datasets.} Median SRCC is reported to avoid target-label fitting. Ridge regularization is selected using source validation data, and the target contributes only test images. (A) Synthetic sources. (B) Authentic sources.}
    \label{tab:authentic_transfer}
    \begin{tabular}{@{}llcc@{}}
            \toprule
            \multicolumn{4}{c}{\textbf{(A) Synthetic sources}} \\
            \textbf{Source} & \textbf{Target} & \textbf{SHAMISA} & \textbf{ARNIQA} \\
            \midrule
            LIVE      & FLIVE & \textbf{0.394} & 0.386 \\
            LIVE      & SPAQ  & \textbf{0.796} & 0.789 \\
            CSIQ      & FLIVE & 0.174 & \textbf{0.222} \\
            CSIQ      & SPAQ  & \textbf{0.639} & 0.520 \\
            TID2013   & FLIVE & \textbf{0.369} & 0.363 \\
            TID2013   & SPAQ  & \textbf{0.640} & 0.619 \\
            KADID-10K & FLIVE & 0.374 & \textbf{0.402} \\
            KADID-10K & SPAQ  & \textbf{0.738} & 0.703 \\
            \midrule
            \multicolumn{4}{c}{\textbf{(B) Authentic sources}} \\
            \textbf{Source} & \textbf{Target} & \textbf{SHAMISA} & \textbf{ARNIQA} \\
            \midrule
            FLIVE & SPAQ  & \textbf{0.844} & 0.817 \\
            SPAQ  & FLIVE & \textbf{0.484} & 0.399 \\
            \bottomrule
    \end{tabular}
\end{table}

\section{Additional t-SNE Visualizations}
\label{app:tsne}

\paragraph{Protocol}
We compute t-SNE~\citep{van2008visualizing} on KADID-10K~\citep{kadid10k} using image-level encoder representations $\mathbf{H}$ before the projector. Features are averaged across crops to obtain 10,125 image-level representations. We first apply PCA to 50 dimensions and then compute a two-dimensional t-SNE embedding with a fixed random seed of 123, perplexity 30, learning rate 200, 2000 iterations, PCA initialization, Euclidean distance, and early exaggeration 12. The same settings and images are used for both models, and no additional feature normalization is applied. These visualizations are qualitative diagnostics and do not affect model selection.

\begin{figure}[t]
    \centering
    \includegraphics[width=\linewidth]{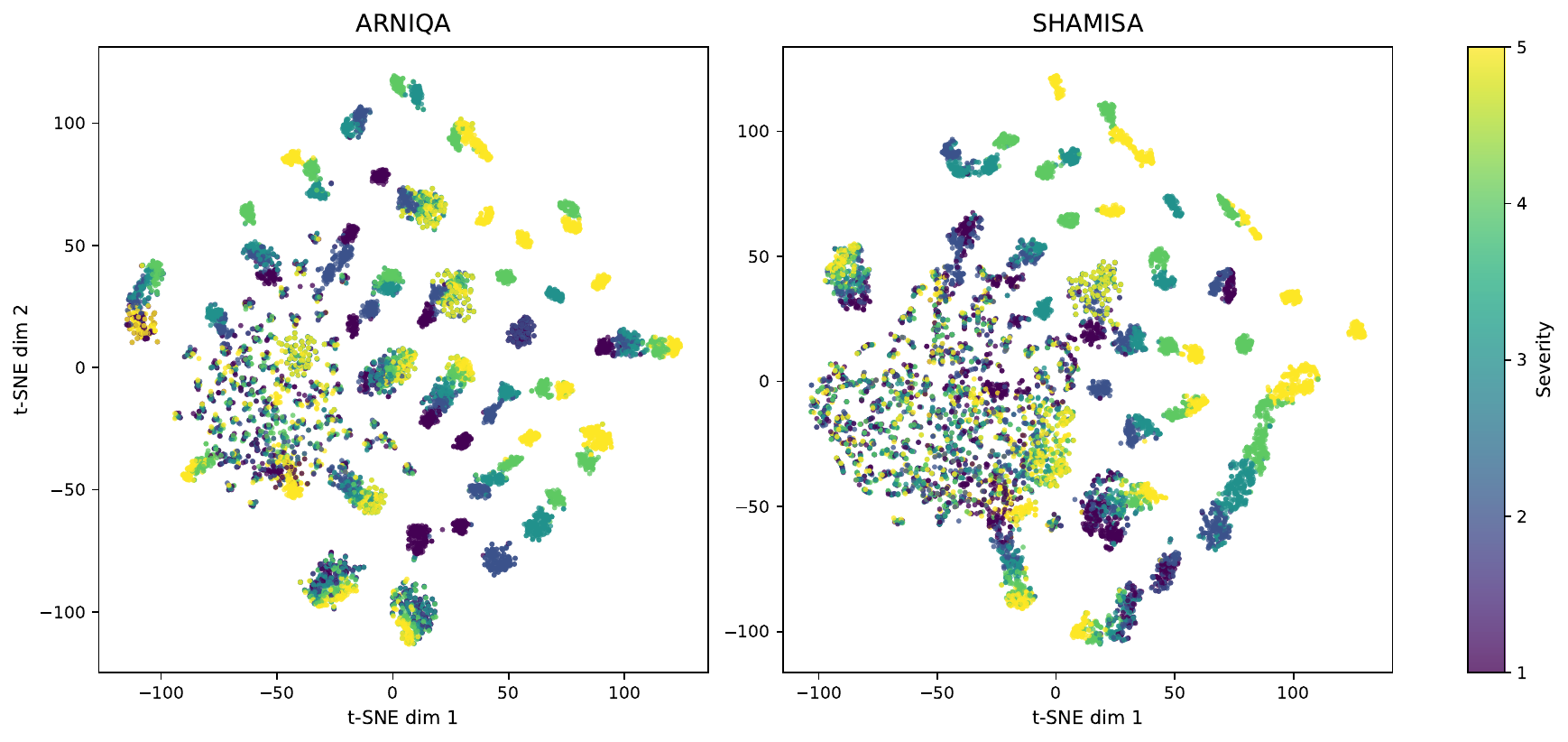}
    \caption{\textbf{t-SNE colored by distortion severity on KADID-10K.} Colors indicate severity levels 1 to 5 for ARNIQA (left) and SHAMISA (right).}
    \label{fig:tsne_kadid_severity}
\end{figure}

\begin{figure*}[t]
    \centering
    \includegraphics[width=\textwidth]{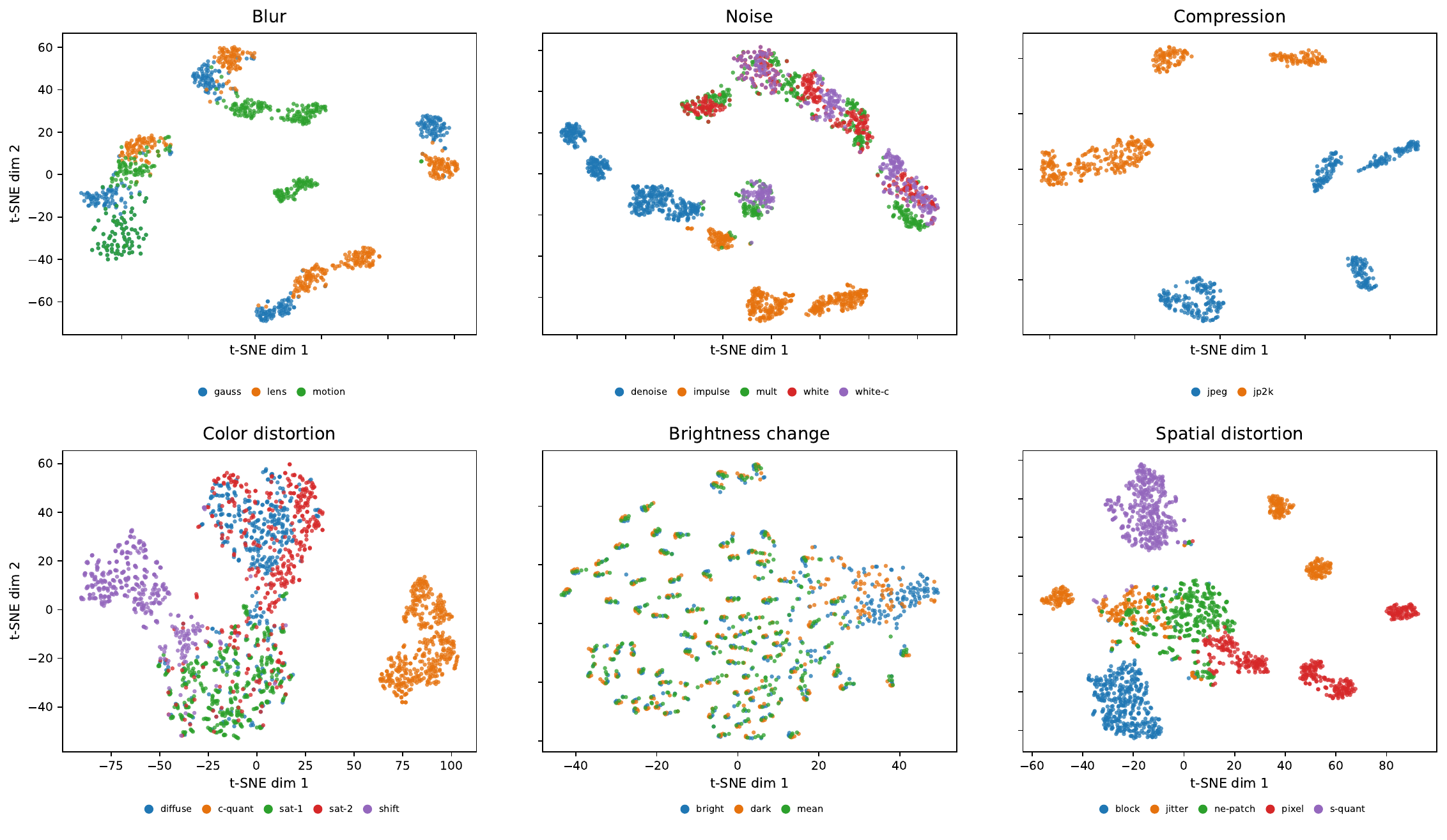}
    \caption{\textbf{Subtype-colored t-SNE views for SHAMISA, including six KADID-10K distortion families.}}
    \label{fig:tsne_kadid_shamisa_subtypes}
\end{figure*}

\paragraph{Additional visualizations}
Fig.~\ref{fig:tsne_kadid_severity} colors the global KADID-10K embedding by severity level and shows smooth local severity progressions within several coarse families, although visually related distortions still overlap. Fig.~\ref{fig:tsne_kadid_shamisa_subtypes} focuses on subtype organization for SHAMISA. Blur, compression, and spatial distortions show the clearest localized subtype regions, while brightness-related and color-related manipulations remain more entangled. These fixed-seed projections provide qualitative support for the learned organization without serving as quantitative separability measures.

\section{UMAP Protocol Details}  
\label{app:umap_protocol}
We sample 1,000 pristine KADIS images and generate blur-only, noise-only, and blur$\rightarrow$noise compositions, with blur applied before noise. The Gaussian blur levels are $\{0.1, 0.5, 1, 2, 5\}$, and the white-noise levels are $\{0.001, 0.002, 0.003, 0.005, 0.01\}$. For each content image, we generate 5 blur-only samples, 5 noise-only samples, and $5 \times 5$ blur$\rightarrow$noise compositions, for a total of 35k images.
We extract $\mathbf{H}$ using the frozen encoder and the feature-pooling protocol in Sec.~\ref{subsec:datasets-protocol}, before the projector head.
We apply no feature-wise standardization or $\ell_2$ normalization to $\mathbf{H}$ before visualization. Features are projected to 50 dimensions with PCA and then embedded with UMAP using 50 neighbors, minimum distance 0.99, Euclidean distance, spectral initialization, and random seed 123. The same settings are used for both models.
For visualization, each point is colored by the relative contributions of blur and noise, and its marker opacity increases with normalized total severity.

\section{gMAD Protocol Details}
\label{app:gmad_protocol}
For each model, we compute scalar quality predictions on the full Waterloo Exploration distorted pool using a frozen encoder and a ridge regressor trained on KADID-10K, following the feature extraction and normalization in Sec.~\ref{subsec:datasets-protocol}.
For a given defender $f_{\mathrm{def}}$ and attacker $f_{\mathrm{att}}$, we bin all images into $N_{\mathrm{bins}}{=}10$ equal-count levels based on $f_{\mathrm{def}}$ predictions and select the lowest and highest bins.
Within each selected bin, we select the top-1 attacker-gap pair, defined as the pair with maximal attacker score separation while remaining in the same defender-defined level, and report it for visualization.
All reported figures use the Waterloo distorted subset.
Table~\ref{tab:gmad_arniqa_trained} summarizes the corresponding top-1 attacker gaps for the four combinations of defender and quality level.

\begin{table}[t]
    \centering
    \small
    \setlength{\tabcolsep}{6pt}
    \caption{\textbf{gMAD summary on the Waterloo Exploration distorted pool for SHAMISA\ and ARNIQA.} Attacker gap denotes the attacker-side score separation of the selected pair within the defender-defined bin. Lower values are better.}
    \label{tab:gmad_arniqa_trained}
    \begin{tabular}{lcc}
        \toprule
        \textbf{Defender, Quality Level} & \textbf{Bin size} & \textbf{Top-1 attacker gap} \\
        \midrule
        ARNIQA, high                     & 9488              & 8.2717                      \\
        ARNIQA, low                      & 9488              & 8.6782                      \\
        SHAMISA, high                    & 9488              & 4.1860                      \\
        SHAMISA, low                     & 9488              & 3.1356                      \\
        \bottomrule
    \end{tabular}
\end{table}

\section{Additional Ablations}
\label{app:ablations}
Table~\ref{tab:ablations_full} reports the complete ablation catalog, including finer-grained graph removals (B2--B5), OT sparsification (C1), the additional distortion-engine settings D3--D4, and severity-dependent edge weighting (F1).
The individual metadata-source removals B3--B5 produce modest changes, while the metadata-only variant B7 remains below A0. This pattern indicates that the metadata relations provide overlapping information and are most useful collectively.

Uniform severity sampling in D3 lowers SRCC$_g$ to 0.8743, while fixing the composition order in D4 lowers it to 0.8587. These results support both the default severity distribution and stochastic composition order.

Removing OT alignment in E1 lowers SRCC$_g$ to 0.7887, supporting the central role of global OT guidance. E2 is a plain VICReg ablation using $25\mathcal{L}_{\mathrm{inv}}+25\mathcal{L}_{\mathrm{var}}+1\mathcal{L}_{\mathrm{cov}}$ and binary invariance pairs between distorted images that share distortion type and severity across contents. It retains the same data, distortion engine, backbone, projector, optimizer, schedule, and training length as SHAMISA. Table~\ref{tab:shared_eval_uncertainty} reports per-dataset medians and sample standard deviations for A0 and E2. FLIVE has one official split, so splitwise standard deviation is not applicable.

Replacing the exponential severity maps with the reciprocal form $1/(1+u)$ in F1 gives SRCC$_g$ of 0.8736. The difference between E1 and F1 suggests that OT guidance provides the stronger contribution, while severity shaping offers an additional refinement.

\begin{table}[t]
    \centering
    \scriptsize
    \setlength{\tabcolsep}{2.6pt}
    \caption{\scriptsize
        \textbf{Complete ablation catalog.}
        Each dataset contributes its median test SRCC and PLCC over the available splits. The synthetic, authentic, and global columns summarize these per-dataset medians over their stated groups. FLIVE uses its single official split.}
    \label{tab:ablations_full}
    \resizebox{\columnwidth}{!}{
        \begin{tabular}{lcccccc}
            \toprule
            \textbf{Variant}                                   &
            \textbf{SRCC$_g$}                                  & \textbf{PLCC$_g$} &
            \textbf{SRCC$_a$}                                  & \textbf{PLCC$_a$} &
            \textbf{SRCC$_s$}                                  & \textbf{PLCC$_s$}                                              \\
            \midrule
            \textbf{A0: SHAMISA}                               & 0.8862            & 0.9042 & 0.7620 & 0.8040 & 0.9483 & 0.9543 \\
            \midrule
            \multicolumn{7}{l}{\textit{(A) Graph aggregation \& regularization}}                                                \\
            A1: Binary $\mG$                                   & 0.8767            & 0.8924 & 0.7475 & 0.7821 & 0.9374 & 0.9457 \\
            A2: No $\mathcal{R}_{\mathrm{graph}}$              & 0.8813            & 0.8981 & 0.7471 & 0.7908 & 0.9450 & 0.9497 \\
            A3: No $\Phi$ (fixed linear mixture)               & 0.8759            & 0.8951 & 0.7479 & 0.7896 & 0.9361 & 0.9457 \\
            A4: No stop-grad through $\Phi$                    & 0.8797            & 0.8980 & 0.7521 & 0.7950 & 0.9395 & 0.9469 \\
            \midrule
            \multicolumn{7}{l}{\textit{(B) Source graph contributions}}                                                         \\
            B1: No $\mG^{o}$ (OT graph)                        & 0.8600            & 0.8787 & 0.7661 & 0.8069 & 0.8995 & 0.9092 \\
            B2: No $\mG^{k}$ (kNN graph)                       & 0.8815            & 0.8989 & 0.7475 & 0.7890 & 0.9452 & 0.9520 \\
            B3: No $\mG^{rd}$ (ref--dist metadata)             & 0.8765            & 0.8931 & 0.7484 & 0.7865 & 0.9366 & 0.9443 \\
            B4: No $\mG^{dd}$ (dist--dist metadata)            & 0.8771            & 0.8937 & 0.7494 & 0.7867 & 0.9370 & 0.9451 \\
            B5: No $\mG^{rr}$ (ref--ref metadata)              & 0.8814            & 0.8976 & 0.7462 & 0.7854 & 0.9458 & 0.9520 \\
            B6: $\mG^{k}$ only                                 & 0.7891            & 0.8018 & 0.7577 & 0.7825 & 0.7922 & 0.8017 \\
            B7: Metadata only $\{\mG^{rd},\mG^{dd},\mG^{rr}\}$ & 0.8467            & 0.8648 & 0.7634 & 0.8010 & 0.8801 & 0.8905 \\
            B8: Structure only $\{\mG^{k},\mG^{o}\}$           & 0.8753            & 0.8961 & 0.7419 & 0.7891 & 0.9387 & 0.9475 \\
            \midrule
            \multicolumn{7}{l}{\textit{(C) Sparsity operators \& topology shaping}}                                             \\
            C1: Row-wise Top-$k$ in $\mG^{o}$ ($k{=}8$)        & 0.8734            & 0.8898 & 0.7500 & 0.7918 & 0.9307 & 0.9359 \\
            \midrule
            \multicolumn{7}{l}{\textit{(D) Distortion engine}}                                                                  \\
            D1: $M_d{=}1$                                      & 0.8775            & 0.9077 & 0.7259 & 0.7684 & 0.9519 & 0.9783 \\
            D2: Discrete severities ($L{=}5$)                  & 0.8767            & 0.8987 & 0.7456 & 0.7890 & 0.9386 & 0.9518 \\
            D3: Uniform severity sampling                      & 0.8743            & 0.8955 & 0.7439 & 0.7946 & 0.9359 & 0.9432 \\
            D4: Fixed composition order $\pi$                  & 0.8587            & 0.8712 & 0.7434 & 0.7668 & 0.9113 & 0.9215 \\
            \midrule
            \multicolumn{7}{l}{\textit{(E) Objective form \& diagnostics}}                                                      \\
            E1: No $\mathcal{L}_{\mathrm{ot}}$                 & 0.7887            & 0.8009 & 0.7597 & 0.7836 & 0.7904 & 0.7996 \\
            E2: Plain VICReg                                  & 0.7942            & 0.8100 & 0.7420 & 0.7555 & 0.8203 & 0.8372 \\
            \midrule
            \multicolumn{7}{l}{\textit{(F) Severity mapping diagnostics}}                                                       \\
            F1: Reciprocal $\phi,\hat{\phi}$                   & 0.8736            & 0.8937 & 0.7459 & 0.7914 & 0.9335 & 0.9423 \\
            \bottomrule
        \end{tabular}
    }
\end{table}

\begin{table}[!t]
    \centering
    \scriptsize
    \setlength{\tabcolsep}{2.4pt}
    \renewcommand{\arraystretch}{1.03}
    \caption{\textbf{Per-dataset split variability.} Entries are median $\pm$ sample standard deviation across repeated splits for A0 and the plain VICReg ablation E2. The dagger ($\dagger$) marks FLIVE results from its single official split, for which splitwise standard deviation is not applicable.}
    \label{tab:shared_eval_uncertainty}
    \resizebox{\columnwidth}{!}{%
    \begin{tabular}{lcccc}
        \toprule
        & \multicolumn{2}{c}{\textbf{A0: SHAMISA}} & \multicolumn{2}{c}{\textbf{E2: plain VICReg}} \\
        \textbf{Dataset} & \textbf{SRCC} & \textbf{PLCC} & \textbf{SRCC} & \textbf{PLCC} \\
        \midrule
        LIVE      & $0.986\!\pm\!0.010$ & $0.987\!\pm\!0.009$ & $0.958\!\pm\!0.011$ & $0.961\!\pm\!0.010$ \\
        CSIQ      & $0.981\!\pm\!0.011$ & $0.987\!\pm\!0.010$ & $0.888\!\pm\!0.024$ & $0.907\!\pm\!0.020$ \\
        TID2013   & $0.904\!\pm\!0.053$ & $0.919\!\pm\!0.099$ & $0.678\!\pm\!0.097$ & $0.723\!\pm\!0.131$ \\
        KADID-10K & $0.922\!\pm\!0.013$ & $0.924\!\pm\!0.013$ & $0.757\!\pm\!0.034$ & $0.758\!\pm\!0.037$ \\
        FLIVE     & $0.610^{\dagger}$      & $0.688^{\dagger}$      & $0.588^{\dagger}$      & $0.610^{\dagger}$ \\
        SPAQ      & $0.914\!\pm\!0.003$ & $0.920\!\pm\!0.003$ & $0.896\!\pm\!0.004$ & $0.901\!\pm\!0.004$ \\
        \bottomrule
    \end{tabular}}
\end{table}

\section{Additional Pre-Training Dynamics} \label{sec:appendix_training_dynamics}

\subsection{Downstream SRCC During SSL Pre-Training} \label{sec:appendix_training_metrics}
We track downstream SRCC of the frozen pre-trained encoder across pre-training snapshots using the same lightweight regressor as in the main experiments.
Fig.~\ref{fig:srcc_vs_steps} complements the main text by showing that SRCC improves rapidly in the early phase and then largely saturates across datasets. The most favorable checkpoint varies because the self-supervised objective is not optimized directly for SRCC on any single benchmark.

\subsection{Encoder Representations and Projector Embeddings During SSL Pre-Training} \label{sec:appendix_rep_diagnostics}
\begin{figure}[t]
    \centering
    \subfloat[\emph{corr} for $\mH$ vs. steps.\label{fig:corr_H}]{%
        \includegraphics[width=0.49\columnwidth]{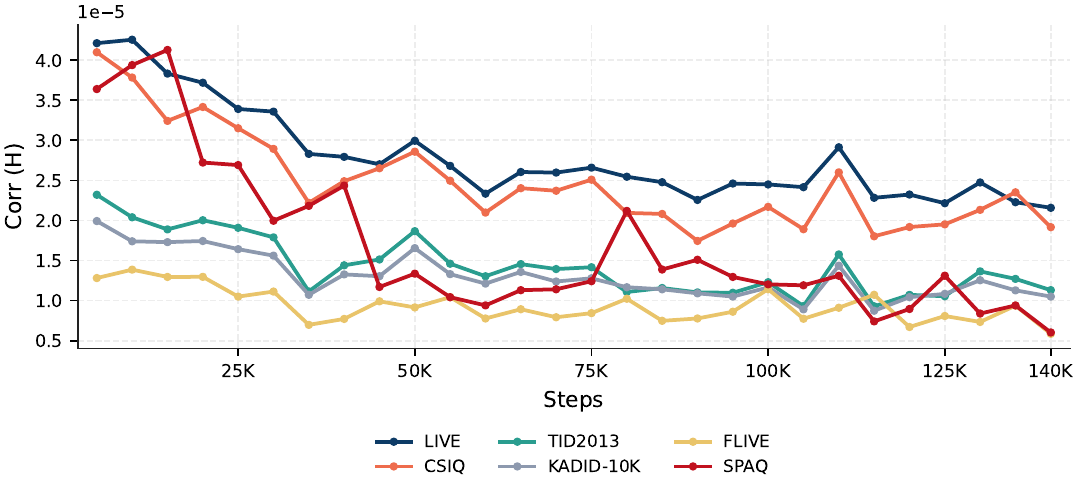}}
    \hfill
    \subfloat[\emph{corr} for $\mZ$ vs. steps.\label{fig:corr_Z}]{%
        \includegraphics[width=0.49\columnwidth]{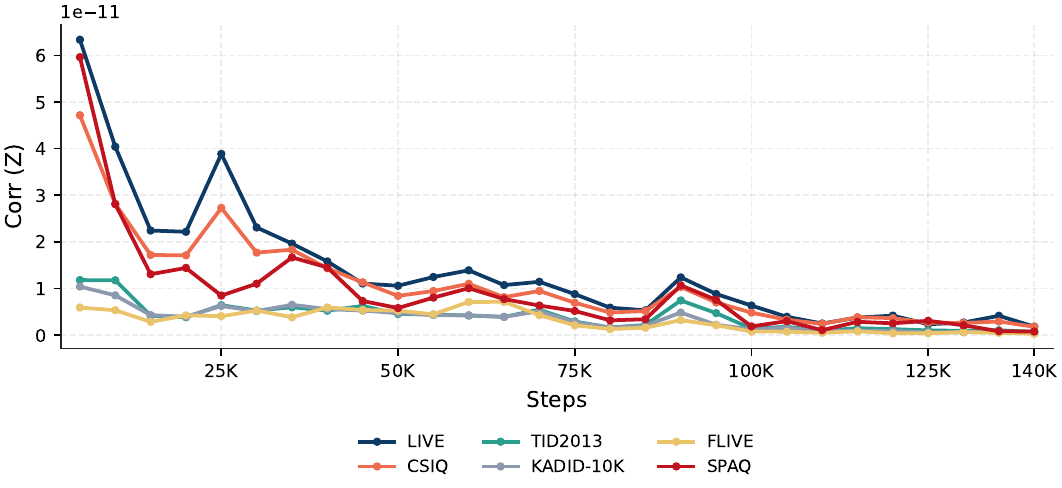}}
    \vspace{\figup}
    \caption{
        \textbf{Feature decorrelation proxy (\emph{corr}) over SSL pre-training.}
        Both $\mH$ and $\mZ$ show a rapid early drop followed by stabilization, consistent with progressively reduced feature redundancy.
    }
    \label{fig:corr_HZ}
    \vspace{\figdown}
\end{figure}

\begin{figure}[t]
    \centering
    \includegraphics[width=0.66\linewidth]{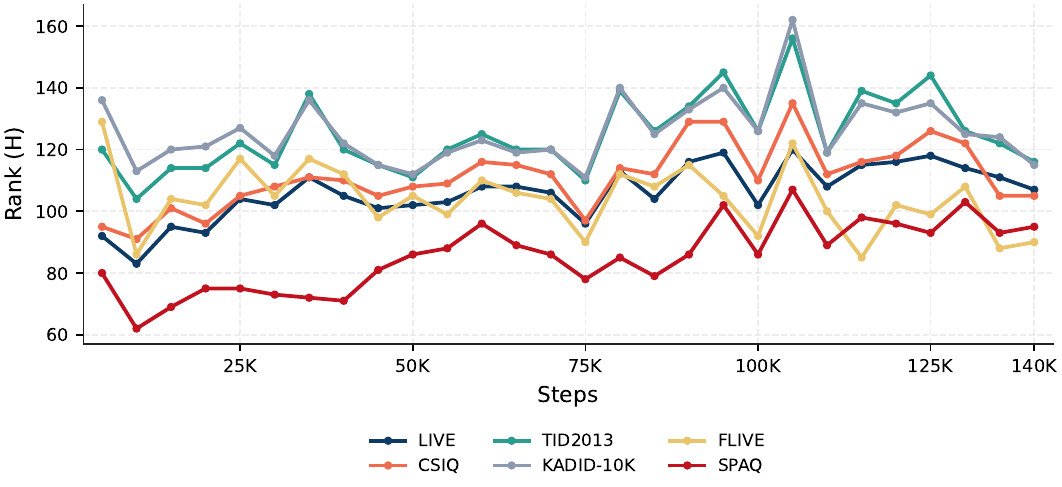}
    \caption{
        \textbf{Effective dimensionality proxy (\emph{rank}) for $\mH$ over SSL pre-training.}
        The encoder representation uses a substantial number of active dimensions throughout training and then stabilizes after the early phase.
    }
    \label{fig:rank_H_only}
    \vspace{\figdown}
\end{figure}

\begin{figure}[t]
    \centering
    \subfloat[\emph{inv} for $\mH$ vs. steps.\label{fig:inv_H}]{%
        \includegraphics[width=0.49\columnwidth]{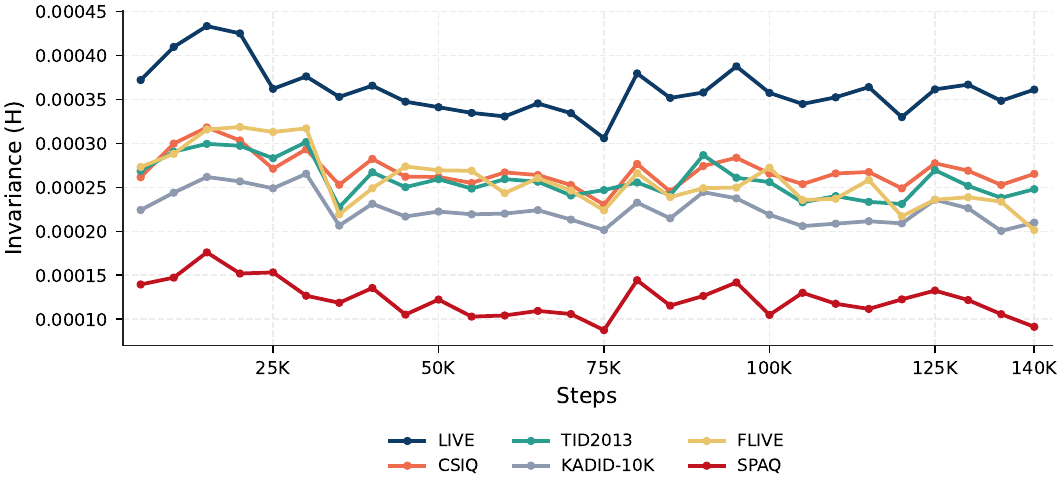}}
    \hfill
    \subfloat[\emph{inv} for $\mZ$ vs. steps.\label{fig:inv_Z}]{%
        \includegraphics[width=0.49\columnwidth]{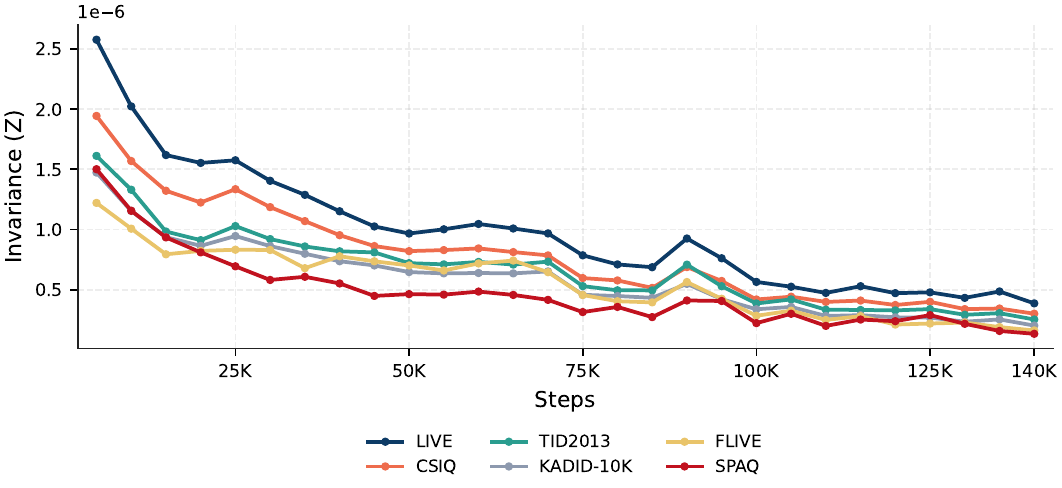}}
    \vspace{\figup}
    \caption{
        \textbf{Invariance proxy (\emph{inv}) over SSL pre-training.}
        Both curves generally decrease and then stabilize, consistent with related samples becoming closer in representation space as training proceeds.
    }
    \label{fig:inv_HZ}
    \vspace{\figdown}
\end{figure}

We additionally track correlation and invariance for both encoder representations $\mH$ and projector embeddings $\mZ$, together with rank for $\mH$, during SSL pre-training.
Fig.~\ref{fig:corr_HZ} shows that the correlation proxy drops quickly and then stabilizes for both $\mH$ and $\mZ$.
Fig.~\ref{fig:rank_H_only} shows that the effective dimensionality of $\mH$ evolves during training and stabilizes after the early phase, consistent with progressively reduced feature redundancy. Fig.~\ref{fig:inv_HZ} shows a similar decrease and stabilization in the invariance proxy, consistent with related samples becoming closer in representation space.

\section{Hyperparameter Sensitivity Details}
\label{sec:hp_sensitivity_appendix}

We perform one-at-a-time sweeps by varying one hyperparameter while keeping all others fixed. Table~\ref{tab:hp_sens_srcc} summarizes SRCC sensitivity on LIVE, CSIQ, TID2013, KADID-10K, FLIVE, and SPAQ. For each hyperparameter, $\Delta$ is the difference between the largest and smallest SRCC across the sweep. Larger $\Delta$ indicates greater sensitivity.

\begin{table}[!t]
    \centering
    \scriptsize
    \setlength{\tabcolsep}{5pt}
    \renewcommand{\arraystretch}{1.05}
    \caption{\textbf{Sensitivity summary for SRCC.} For each hyperparameter, $\Delta$ is the largest minus smallest SRCC across the sweep. Larger $\Delta$ indicates higher sensitivity. The most sensitive and second most sensitive hyperparameters per dataset are shown in \textbf{bold} and \underline{underlined}.}
    \label{tab:hp_sens_srcc}
    \resizebox{\columnwidth}{!}{
        \begin{tabular}{lccccccc}
            \toprule
            HP       & LIVE              & CSIQ              & TID2013           & KADID-10K         & FLIVE             & SPAQ              & Mean              \\
            \midrule
            $d_h$    & 0.012             & 0.003             & 0.003             & 0.006             & 0.004             & \textbf{0.007}    & 0.006             \\
            $d_z$    & 0.006             & 0.002             & 0.002             & 0.000             & 0.003             & \underline{0.007} & 0.003             \\
            \midrule
            $B$      & 0.002             & 0.001             & 0.003             & 0.005             & 0.001             & 0.001             & 0.002             \\
            \midrule
            $k_n$    & 0.007             & 0.009             & 0.006             & 0.016             & 0.007             & 0.001             & 0.008             \\
            $K$      & \underline{0.019} & 0.035             & 0.061             & 0.030             & 0.007             & 0.005             & 0.026             \\
            \midrule
            $\gamma$ & 0.017             & \underline{0.045} & \underline{0.089} & \underline{0.070} & \underline{0.033} & 0.001             & \underline{0.042} \\
            $\alpha$ & 0.003             & 0.004             & 0.006             & 0.010             & 0.003             & 0.001             & 0.004             \\
            $\beta$  & 0.005             & 0.004             & 0.006             & 0.005             & 0.007             & 0.003             & 0.005             \\
            \midrule
            $\xi$    & 0.010             & 0.002             & 0.003             & 0.013             & 0.000             & 0.000             & 0.005             \\
            $\eta$   & \textbf{0.054}    & \textbf{0.096}    & \textbf{0.208}    & \textbf{0.226}    & \textbf{0.035}    & 0.001             & \textbf{0.104}    \\
            \bottomrule
        \end{tabular}
    }
\end{table}